\documentclass[lettersize,journal]{IEEEtran}
\usepackage{mathrsfs}
\usepackage{color}
\usepackage{setspace}

\usepackage{amsmath,amsfonts}
\usepackage{algorithmic}
\usepackage{algorithm}
\usepackage{array}
\usepackage[caption=false,font=normalsize,labelfont=sf,textfont=sf]{subfig}
\usepackage{textcomp}
\usepackage{stfloats}
\usepackage{url}
\usepackage{verbatim}
\usepackage{graphicx}
\usepackage{cite}
\usepackage{times}
\usepackage{bm}

\usepackage{fancyhdr} 
\usepackage{lastpage} 
\usepackage{extramarks} 
\usepackage{graphicx} 
\usepackage{lipsum} 
\usepackage{authblk}
\usepackage{multirow}
\usepackage{multicol}
\usepackage{mathrsfs}
\usepackage{arydshln}
\usepackage{mathtools}
\usepackage{framed}
\usepackage{booktabs,multirow}
\usepackage{xcolor}
\usepackage{listings}
\usepackage{textcomp}
\usepackage{setspace}
\usepackage{palatino}
\usepackage{pdflscape}
\usepackage{soul}

\newcommand{\fb}{\mathbf{f}}

\newcommand{\x}{\mathbf{x}}

\newcommand{\X}{\mathbf{X}}
\newcommand{\A}{\mathbf{A}}
\newcommand{\C}{\mathbf{C}}

\newcommand{\w}{\mathbf{w}}
\newcommand{\br}{\mathbf{r}}

\newcommand{\bv}{\mathbf{v}}
\newcommand{\bg}{\mathbf{g}}
\newcommand{\bV}{\mathbf{V}}
\newcommand{\bK}{\mathbf{K}}

\newcommand{\bR}{\mathbf{R}}
\newcommand{\bG}{\mathbf{G}}
\newcommand{\bP}{\mathbf{P}}
\newcommand{\bQ}{\mathbf{Q}}

\newcommand{\Y}{\mathbf{Y}}
\newcommand{\bI}{\mathbf{I}}

\newcommand{\bSigma}{\boldsymbol{\Sigma}}
\newcommand{\bOmega}{\boldsymbol{\Omega}}

\newcommand{\E}{\mathrm{E}}
\newcommand{\Cov}{\mathrm{Cov}}
\newcommand{\calR}{\mathcal{R}}

\newcommand{\calZ}{\mathcal{Z}}

\newcommand{\I}{\mathbf{I}}

\newcommand{\bH}{\mathbf{H}}
\newcommand{\bLambda}{\mathbf{\Lambda}}

\newtheorem{theorem}{\textbf{Theorem}}
\newtheorem{lemma}{\textbf{Lemma}}
\newtheorem{proposition}{\textbf{Proposition}}

\newtheorem{assumption}{\textbf{Assumption}}

\newtheorem{remark}{\textbf{Remark}}

\begin{document}

\title{Partially-Observable Sequential Change-Point Detection for Autocorrelated Data via Upper Confidence Region}

\author{Haijie Xu$^{1}$, Xiaochen Xian$^{2}$, Chen Zhang$^{1}$,  Kaibo Liu$^{3}$
\thanks{$^{1}$Tsinghua University, Beijing, China}
\thanks{$^{2}$University of Florida, Gainesville, FL}
\thanks{$^{3}$University of Wisconsin-Madison, Madison, WI}}

\maketitle






\begin{abstract}
Sequential change point detection for multivariate autocorrelated data is a very common problem in practice. However, when the sensing resources are limited, only a subset of variables from the multivariate system can be observed at each sensing time point. This raises the problem of partially observable multi-sensor sequential change point detection. For it, we propose a detection scheme called adaptive upper confidence region with state space model (AUCRSS). It models multivariate time series via a state space model (SSM), and uses an adaptive sampling policy for efficient change point detection and localization. A partially-observable Kalman filter algorithm is developed for online inference of SSM, and accordingly, a change point detection scheme based on a generalized likelihood ratio test is developed. How its detection power relates to the adaptive sampling strategy is analyzed. Meanwhile, by treating the detection power as a reward, its connection with the online combinatorial multi-armed bandit (CMAB) problem is formulated and an adaptive upper confidence region algorithm is proposed for adaptive sampling policy design. Theoretical analysis of the asymptotic average detection delay is performed, and thorough numerical studies with synthetic data and real-world data are conducted to demonstrate the effectiveness of our method.
\end{abstract}


\section{Introduction}
\label{sec:intro}
Unsupervised learning for sequential change-point of multivariate data streams is ubiquitous in a wide range of applications. The main objective is to detect assignable causes as soon as they occur while maintaining a certain global false alarm rate. 
For instance, in manufacturing, it enables practitioners to identify abnormal changes in machine conditions for quality control. Similarly, in transportation management, it aids in detecting irregular traffic flows, such as congestion at various locations. Additionally, in cybersecurity, it facilitates the monitoring of networks to identify malicious activities. These scenarios typically involve numerous variables that capture real-time information about the system's state.
Real-time monitoring of these variables can serve for efficient in-time change point detection.

However, for many real-world problems, continuous observation and monitoring of such a large number of variables may be infeasible. Instead, only a subset of the variables can be observed for monitoring, due to certain sensing resource constraints, such as the following scenarios: (1) When the cost of sensors is very high, we can not afford sensors for measuring all variables. (2) When the batteries or lifetime of sensors are limited and sensor replacement is difficult, we can not put all sensors in ``ON'' mode. (3) Even though all variables can be sensed in high resolution, this high-dimensional data may not be able to be fully transmitted to the data analysis center due to the limited transmission bandwidth. Consequently, we can only obtain a subset of the observations for real-time monitoring. These partially-observable scenarios lead to information loss and consequently raise demand for monitoring schemes with more accurate estimation and detection power. Furthermore, these partially observable scenarios additionally require monitoring schemes to determine which variables to observe adaptively, such that the observed data can provide the most useful information about change in the system. 

\begin{figure*}
    \centering
    \subfloat[]{

            \includegraphics[scale = 0.11]{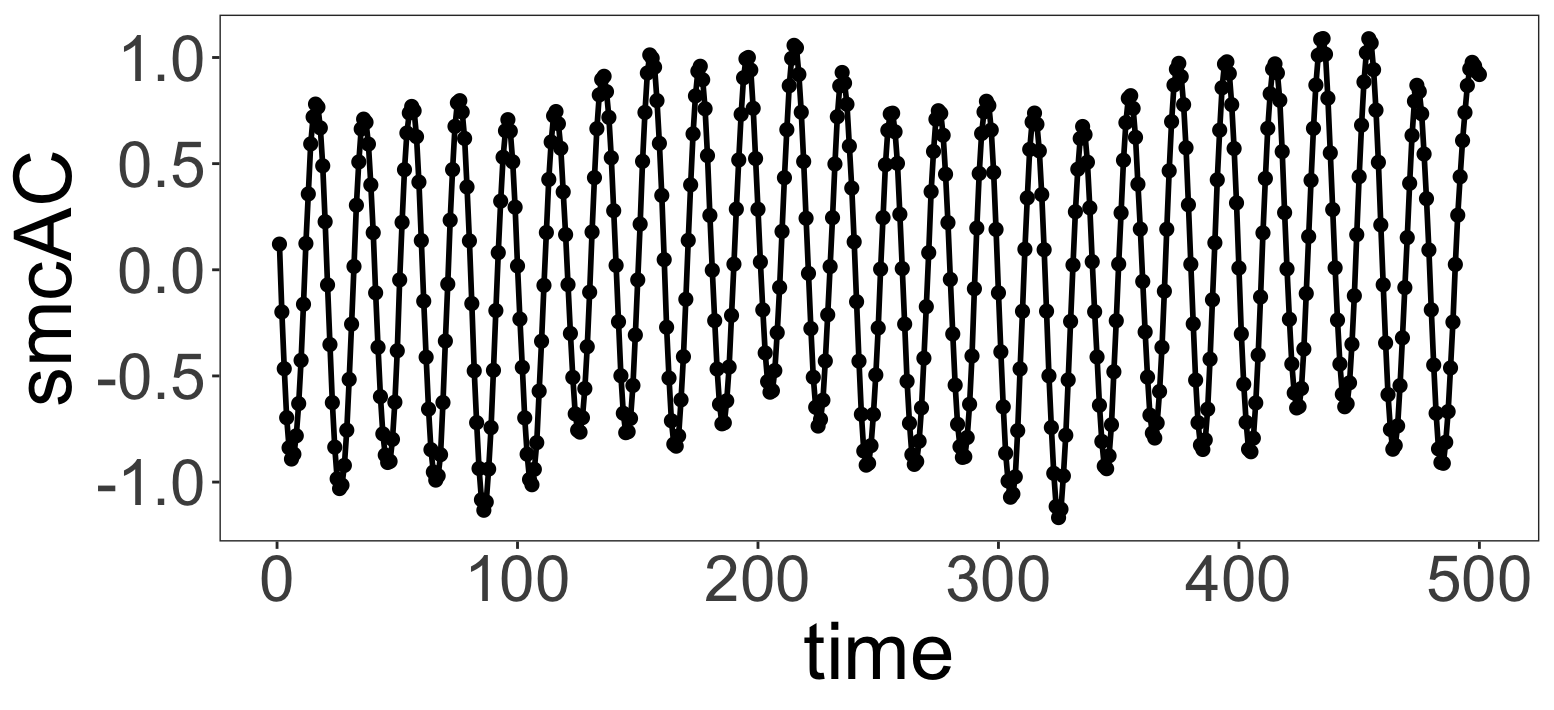}
            \label{fig:case data 1}
        
        }
    \subfloat[]{
        
            \includegraphics[scale = 0.11]{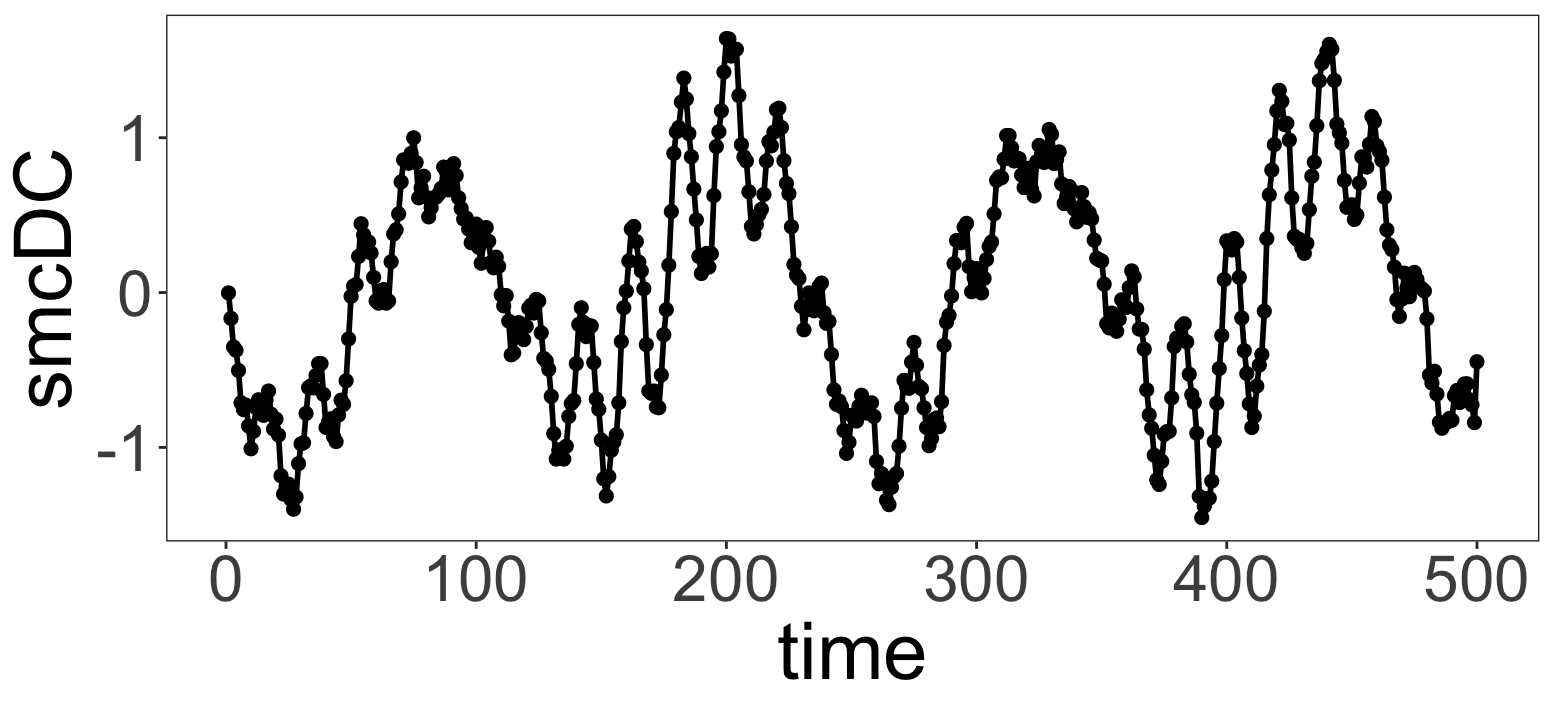}
            \label{fig:case data 2}
        }
        
    \subfloat[]{
       
            \includegraphics[scale = 0.11]{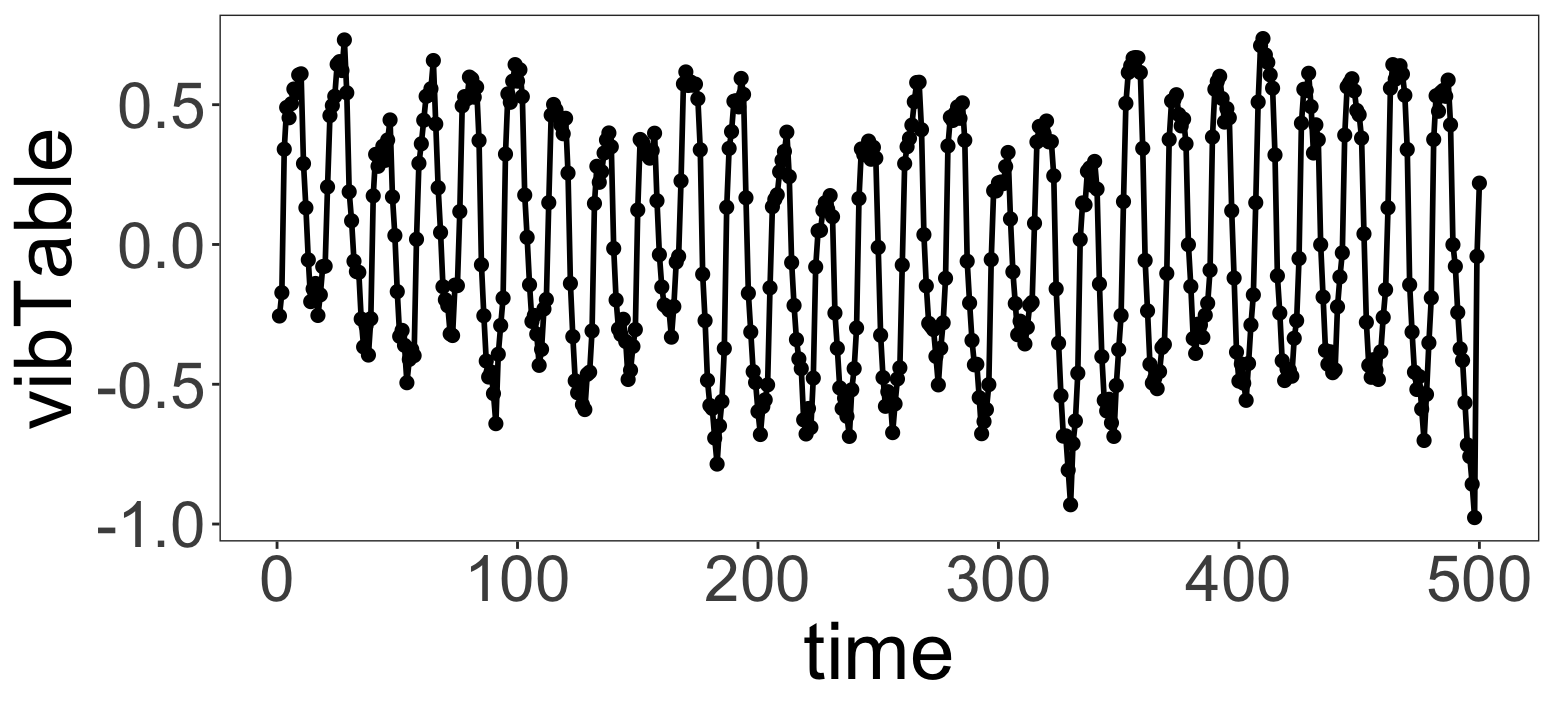}
            \label{fig:case data 3}
        }
    \subfloat[]{
        
            \includegraphics[scale = 0.11]{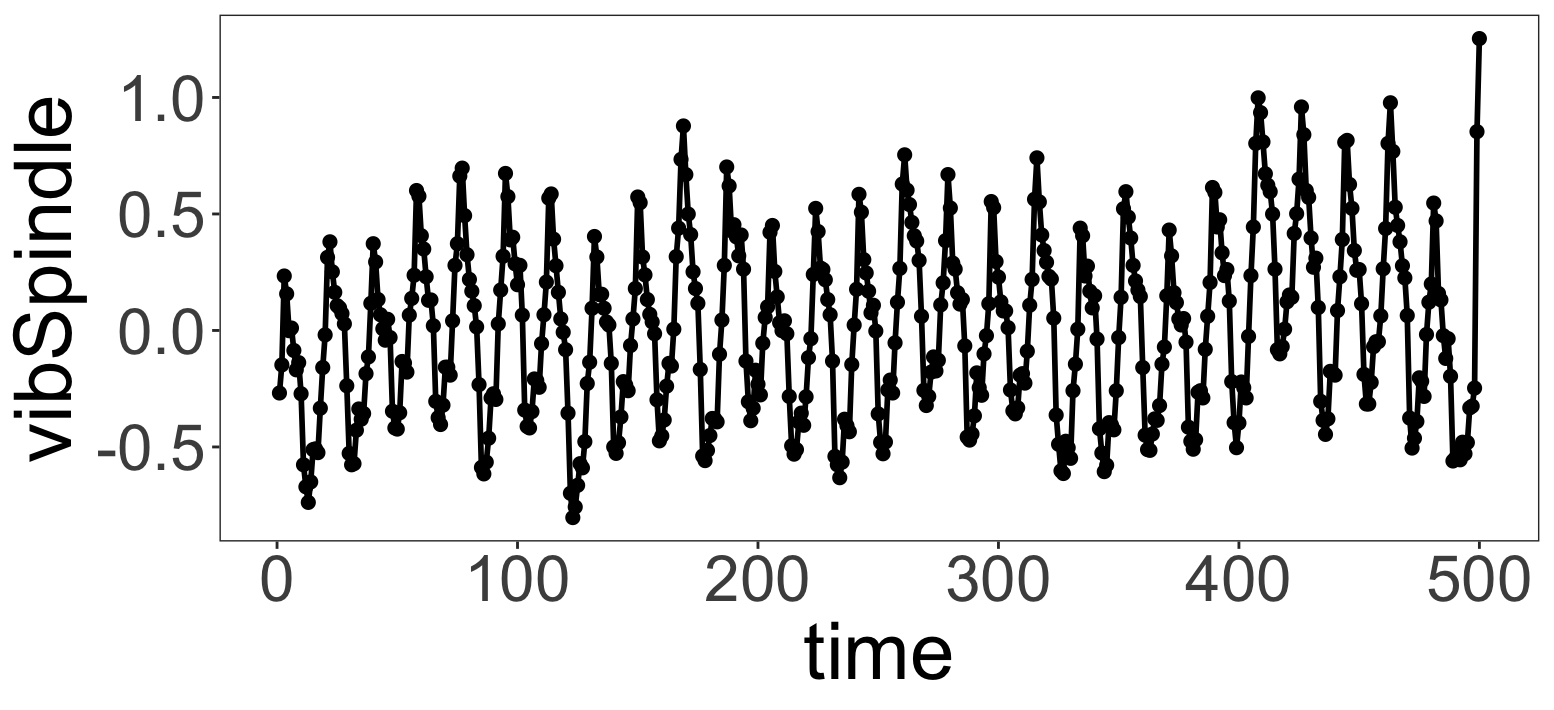}
            \label{fig:case data 4}
        }
        
    \subfloat[]{
            \includegraphics[scale = 0.11]{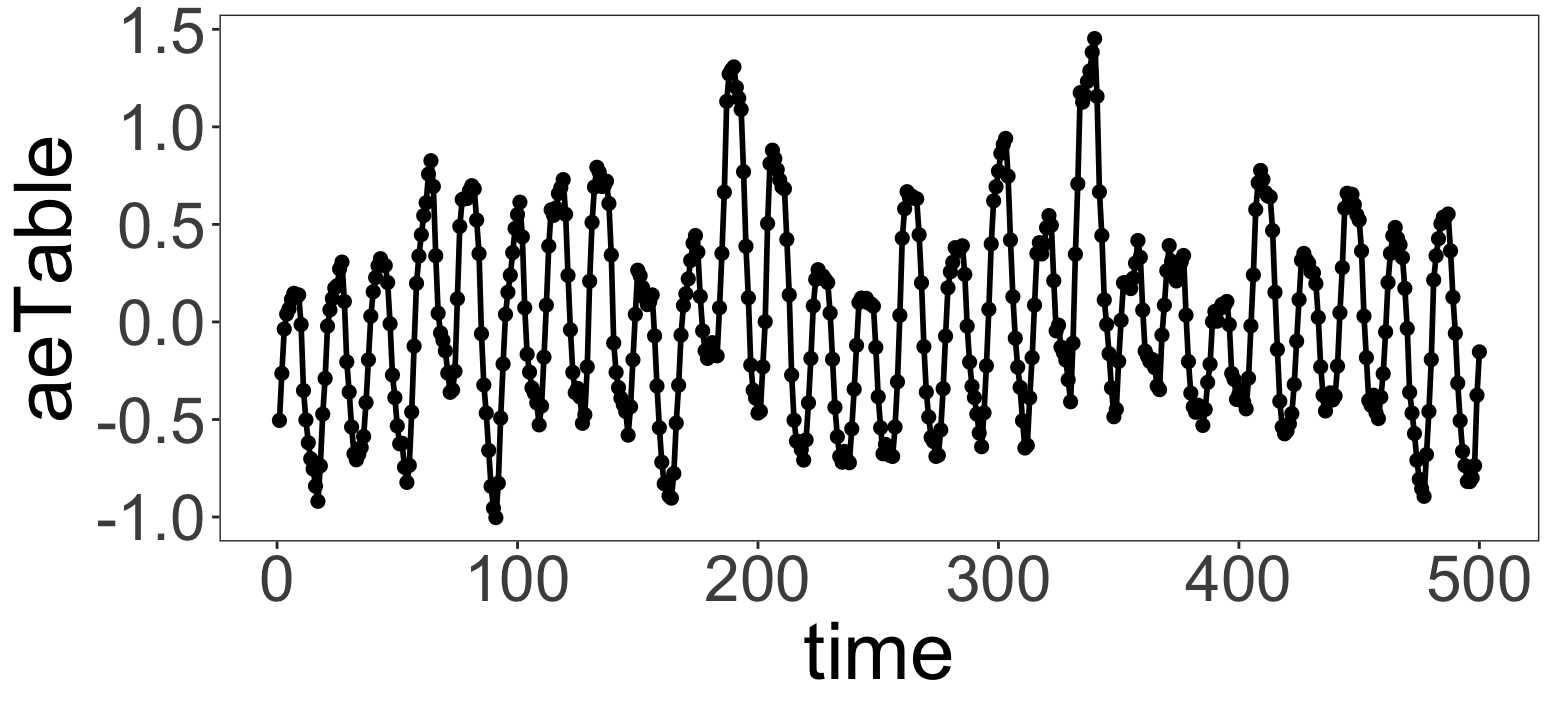}
            \label{fig:case data 5}
        }  
    \subfloat[]{
            \includegraphics[scale = 0.11]{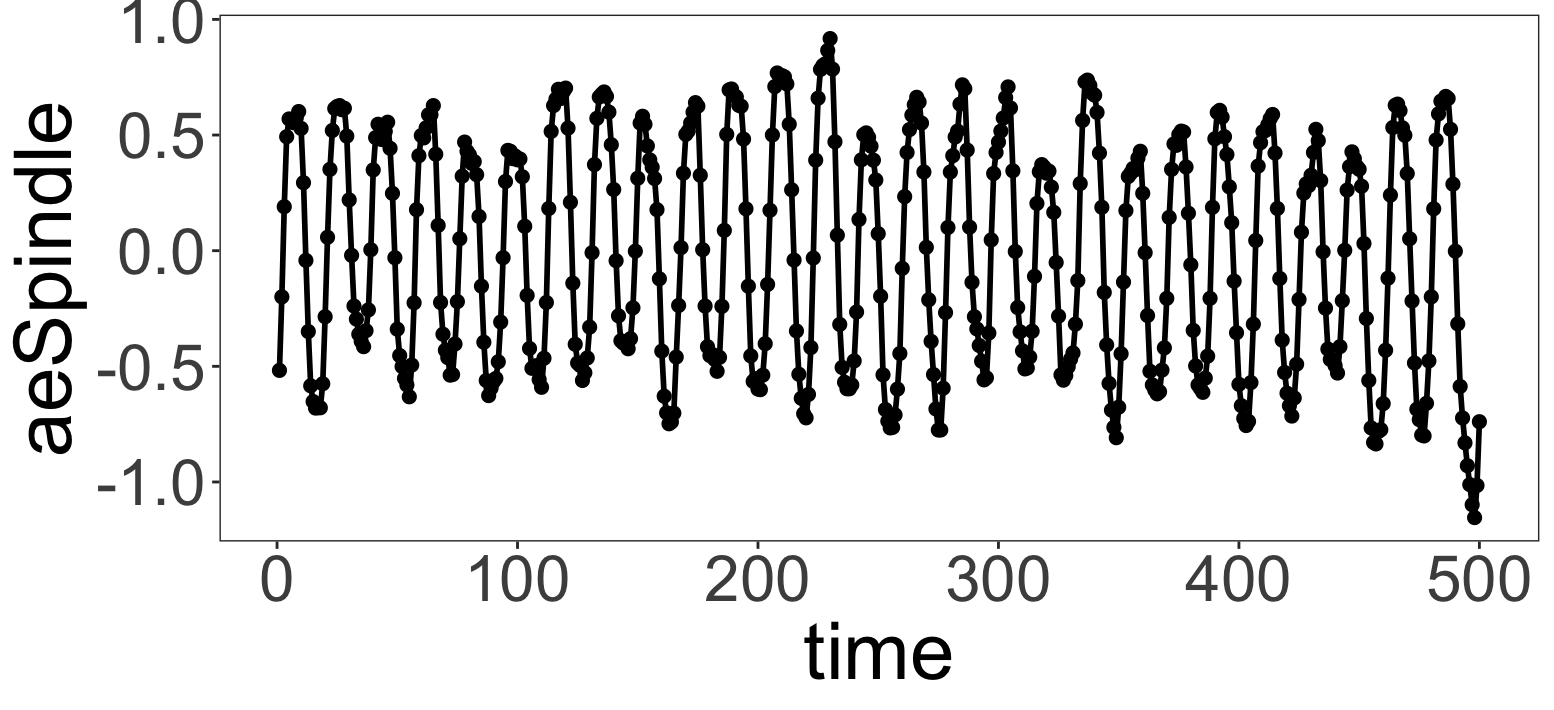}
            \label{fig:case data 6}
        }
    \caption{Stream data of six variables (a)-(f) over 500 time points in a milling process}
    \label{fig:case data in six}
\end{figure*}

\begin{figure*}[htb]
    \centering
    \subfloat[]{
    \includegraphics[scale = 0.325]{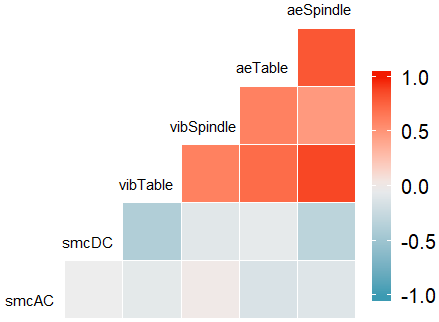}
            \label{fig:intro_corre}
    }
    \subfloat[]{
    \begin{minipage}[b]{0.6\linewidth}
    \includegraphics[scale = 0.094]{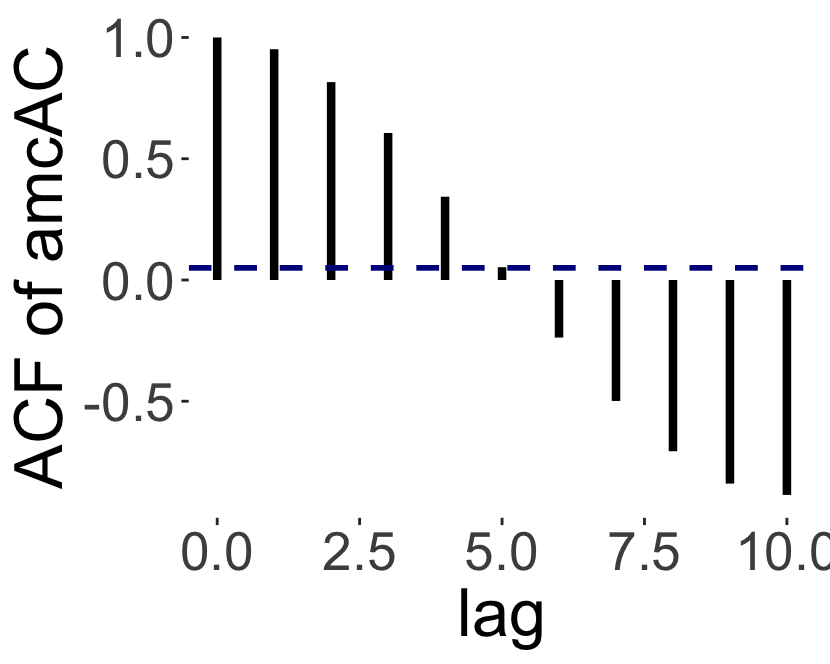}
            \label{fig:acf case 1}
    \includegraphics[scale = 0.094]{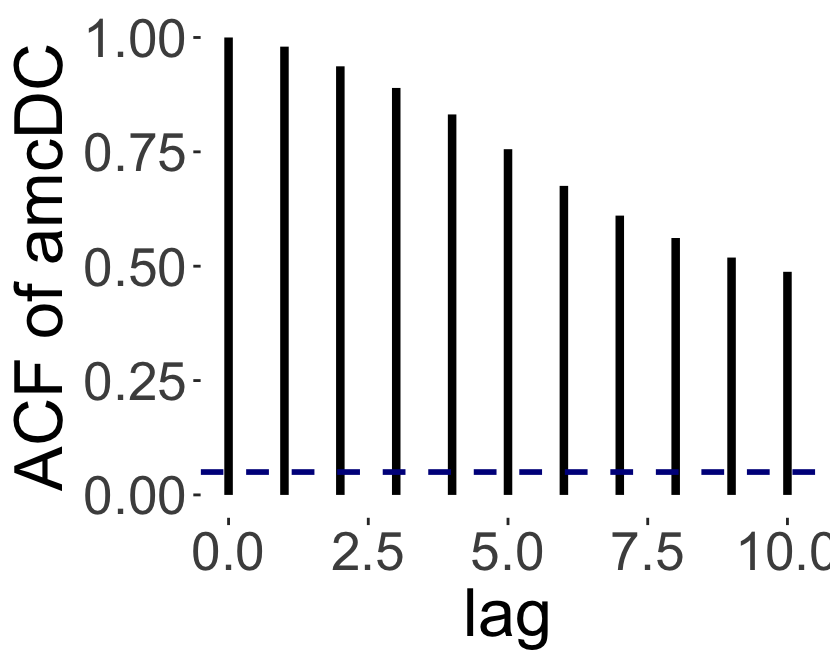}
            \label{fig:acf case 2}    
    \includegraphics[scale = 0.094]{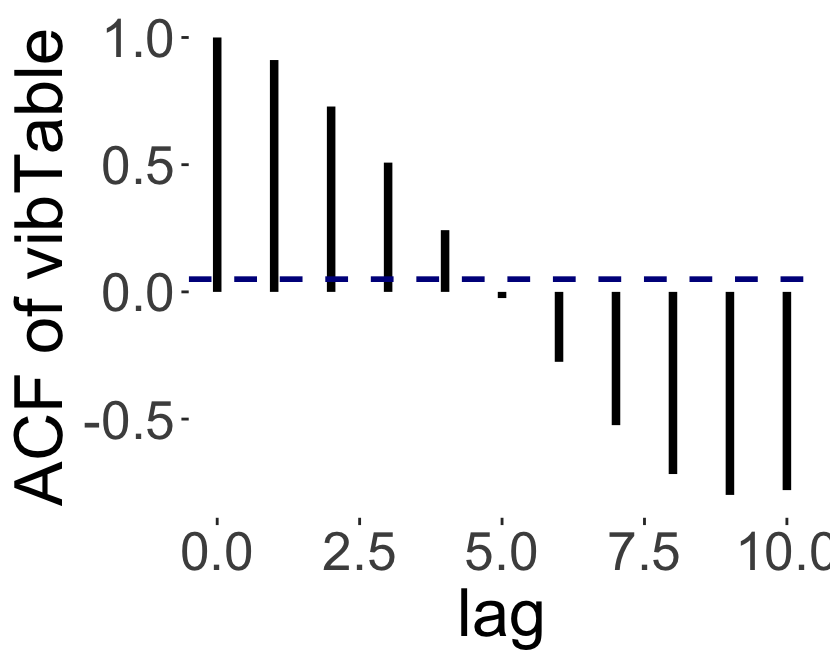}
            \label{fig:acf case 3} 
            
    \includegraphics[scale = 0.094]{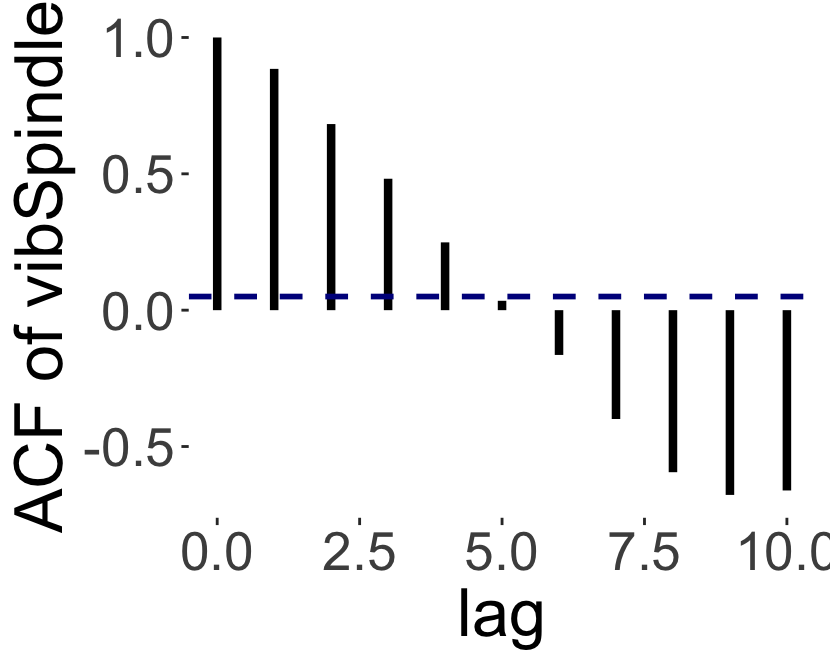}
            \label{fig:acf case 4}
    \includegraphics[scale = 0.094]{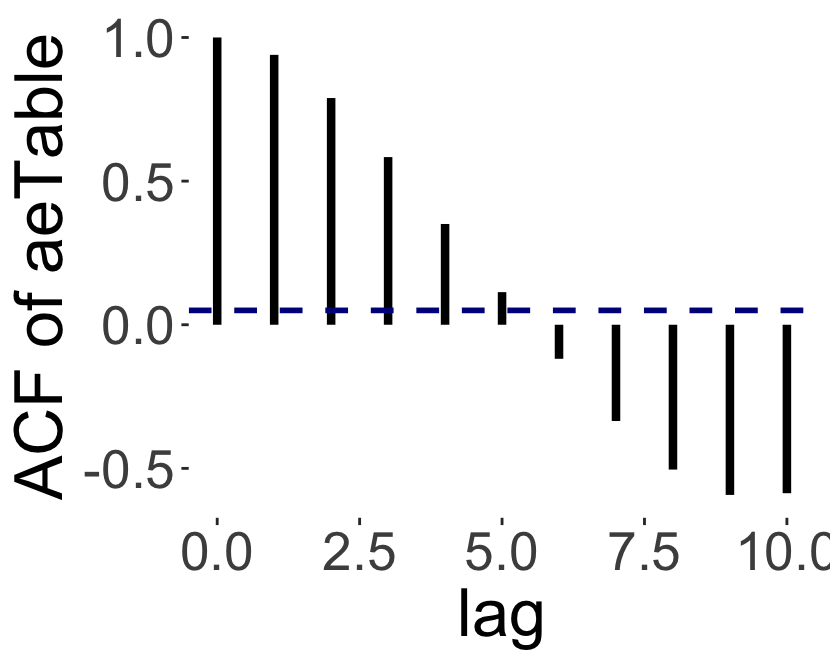}
            \label{fig:acf case 5}
    \includegraphics[scale = 0.094]{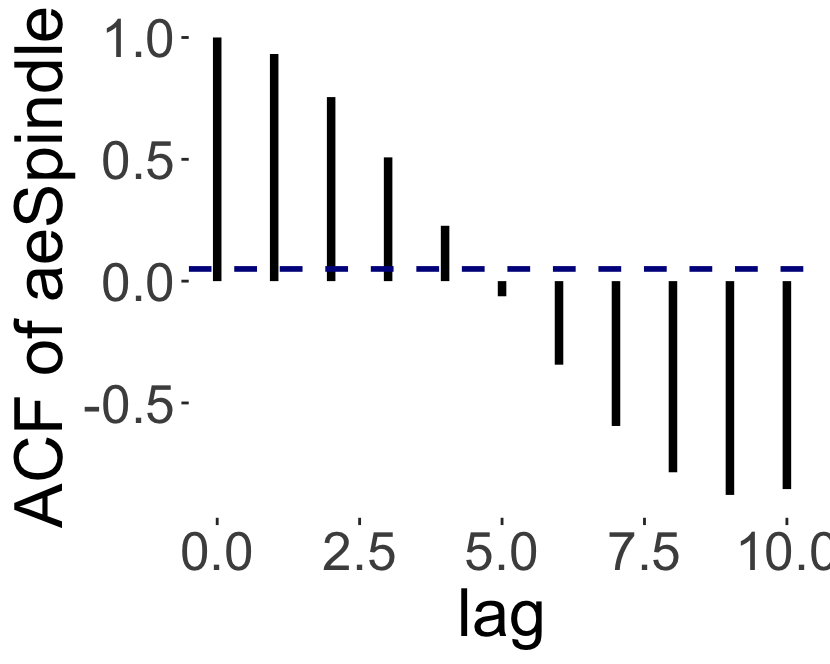}
            \label{fig:acf case 6}
    \end{minipage}
    \label{fig:acf of case data}
    }
    \caption{Person correlation (a) and autocorrelation function (ACF) (b) of  data of milling process}
    \label{fig:corr and acf}
\end{figure*}

One characteristic of multivariate data stream is that different variables exhibit complex cross-correlations. Figure \ref{fig:case data in six} shows sequentially collected data from six variables in a milling process, including AC spindle motor current (smcAC), DC spindle motor current (smcDC), table vibration (vibTable), spindle vibration (vibSpindle), acoustic emission at table (aeTable) and acoustic emission at spindle (aeSpindle). The correlations of these variables are shown in Figure \ref{fig:intro_corre}. We can clearly find that the first four variables have strong positive correlations with each other and have strong negative correlations with the rest two. This indicates it is necessary to consider their correlations in detection. On the one hand, ignoring these correlations may lose information and sacrifice detection power. On the other hand, by considering correlations between variables, even though a variable is not observed at a time point, its information can be reflected by other observed variables that correlate with it. In this way, the information can be better inferred. 

Another characteristic of streaming data is that observations at sequential time points are usually autocorrelated, especially when the sensing frequency is high. Figure \ref{fig:acf of case data} verifies this point by showing the autocorrelation functions of the six variables in the milling process. This means that the change information will accumulate over time. Furthermore, it may transit from one variable to another. As such, no consideration of autocorrelations of data stream will cause a loss of detection power. 

So far online monitoring for partially-observable multivariate data streams has attracted increasing research interest, since the most cornerstone work of \cite{liu2015adaptive}. Most pioneer works do not consider the cross-correlation of different variables \cite{wang2018spatial,xian2021online}. They constructed detection schemes for each variable separately and then fused the results together for system monitoring. Later works take cross-correlation of variables into account for better detection power achievement \cite{zhang2019partially,xian2018nonparametric}. The above works consider multivariate processes with normal distributions. Some other extensions consider more general non-normal distributions, and propose some non-parametric monitoring schemes. A detailed literature review of the current methods is shown in Section \ref{sec:related work}. However, so far to our best knowledge, no work can deal with autocorrelation of data streams except \cite{gomez2021adaptive}, where it uses tensor decomposition to deal with the correlation in the time dimension, i.e., autocorrelation, and further assumes the autocorrelation can change over time smoothly. The anomaly is defined as abrupt change that cannot be described by the smooth tensor decomposition. This setting is different from ours. Actually, when the anomaly is not so abrupt, this method is very likely to treat it as a smooth background and misdetect it. as illustrated in Section \ref{sec:numeric study}.

For multivariate autocorrelated data monitoring, so far tons of works have been proposed. They generally first construct a time series model for data, such as autoregressive and moving average models (ARMA) \cite{jiang2000new}, state-space models (SSM) \cite{pan2004applying, shang2013statistical}, and then construct monitoring statistic to detect the model changes, which can be reflected by the increased fitting error \cite{apley1999glrt, pan2004applying}, or the change of model parameters \cite{chen2011multivariate}. However, extending these methods to partial-observations is not trivial, which increases much difficulty for the time series model inference algorithm. Furthermore, since the abnormal dimensions are usually few, to ensure these changed dimensions can be observed, the adaptive sampling policy should be carefully designed together with the monitoring scheme.  

The topic of sequential adaptive sampling for partially-observable data streams is highly related to the combinatorial multi-armed bandit (CMAB) problem. In CMAB, the bandit faces with several arms with unknown reward distributions. These arms can be either correlated or not. Every time the bandit selects a certain number of arms, which are defined as a combinatorial super arm, and will return a joint reward. Based on the collected rewards over time, the bandit can learn the reward distributions of each super arm better. The goal of CMAB is to maximize the total reward over a certain time horizon. The CMAB problem is very similar to our problem, where each super arm is a combination of selected variables.
For each super arm, we can propose a hypothesis test, and the larger the hypothesis test statistic is, the more likely the super arm is to be abnormal. Then we can use the statistic as the super arm's reward which can measure how much information about the anomaly selecting this super arm can give us. 
Similar to CMAB, we have no knowledge about which variable has changed. On the one hand, we may choose those variables which, based on their past observations, are estimated to be the most changed ones. This is the so-called exploitation. On the other hand, we may choose to observe the variables which have not been observed enough times, since they still have chance to be the most changed ones with more observations. This is the so-called exploration. Sampling methods considering exploitation and exploration have been thoroughly designed for CMAB, such as the sampling based on upper confidence bound \cite{carpentier2011upper}  and Thompson sampling \cite{ye2022online}. They provide potentially insightful solutions to our problem, as reviewed in Section \ref{sec:related work}.

In this paper, we propose a novel method for partially-observable multivariate data stream change point detection (POMDSCD). Our key contributions are four-fold: (i) We formulate multivariate streaming data via an SSM, and propose a partially-observable Kalman filter algorithm for model inference; (ii) Accordingly, we construct a change point detection scheme based on general likelihood ratio test (GLRT) and analyze how its detection power is related to the adaptive sampling strategy; (iii) By treating the detection power as a reward function for each selected subset of variables, we formulate our problem as a CMAB and propose an adaptive upper confidence region algorithm for sampling policy design; (iv) Last, we give theoretical analyses of our proposed method's detection power.

The remainder of this paper is organized as follows. Section \ref{sec:related work} reviews the literature on related topics to our problem. Section \ref{sec:formulation} describes our specific problem formulation. Section \ref{sec:methodology} introduces our proposed method together with its theoretical properties in detail. Section \ref{sec:numeric study} presents thorough numerical studies based on simulated data. Section \ref{sec:case study} applies our proposed method to a real-world case study to further illustrate its efficacy and applicability. Some conclusive remarks are given in Section \ref{sec:conclusion}.

\section{Related Work}
\label{sec:related work}
\subsection{Online monitoring for data streams}

Online monitoring for data streams has been extensively studied in the literature of statistics, machine learning and beyond \cite{xie2013sequential,tartakovsky2014sequential,chan2017optimal,aminikhanghahi2017survey}. Most of the existing works are under the change-point model formulation: $\Y_{t} \sim f_{0}, t \leq \tau$ and $\Y_{t}  \sim f_{1}, t > \tau $, where $\tau$ is the unknown change point to be detected. Most pioneer works assume $\Y_{t}$ are identically and independently distributed for different $t$, and propose various types of detection statistics \cite{wang2018spatial,liu2015adaptive}. Among them, methods based on the generalized likelihood ratio test (GLRT) prove to be one of the most robust and efficient \cite{maillard2019sequential,zou2008directional}.

As the systems to be monitored become more complex and data sensing frequency increases, sequentially collected data begin to show strong autocorrelations,
and monitoring methods for autocorrelated data become more popular. In particular, for one-dimensional time series data, ARMA models were first applied to fit the data, and accordingly GLRTs were used to construct monitoring statistics \cite{apley1999glrt, jiang2000new}. To better capture the dynamic evolution of data streams, especially from manufacturing systems, SSM was introduced for process modeling and monitoring   \cite{shi1999state,shi2012.}. Compared with ARMA models, SSM assumes that the sensed data are realizations of some system's state variables, which are latent and evolve according to a first-order AR model. With this two-layer construction, SSM can describe the process autocorrelations more flexibly and have particular applicability and efficiency for change point detection of manufacturing systems.  
In particular, \cite{kawahara2007change} used SSM  to fit multivariate time series data and applied the subspace identification method to find the change point. \cite{pan2004applying} used SSM to fit the multivariate autocorrelated data and conducted Hotelling $T^2$ for change point detection. 
Besides these, there are also many methods that can monitor other different kinds of data streams such as multivariate count data \cite{das2016statistical} , categorical data \cite{das2017detecting} or Gaussian process \cite{kontar2017estimation}.
However, all these methods mentioned above assume that  $\Y_t$ should be fully observable and require nontrival efforts to be extended to POMDSCD cases.

\subsection{Partially-observable multivariate data stream change point detection (POMDSCD)}
Until the most recent ten years, as the demand for advanced system monitoring with big data increased, the problem of POMDSCD began to rise up attentions. \cite{liu2015adaptive} proposed a top-R detection scheme by extending \cite{mei2010efficient} to the POMDSCD scenario. It treats different variables as independent without exploiting their correlation structure.
Later, \cite{wang2018spatial,xian2021online} proposed spatial-adaptive sampling and monitoring procedures by utilizing the correlation of multivariate data. However, all of these adaptive sampling strategies are heuristic, and their adaptive sampling strategies are based on a rule of thumb. \cite{zhang2019partially,nabhan2021correlation} exploited the relationship between POMDSCD and CMAB, and proposed adaptive sampling strategies based on the upper confidence bound (UCB). \cite{nabhan2021correlation} also demonstrated that  \cite{liu2015adaptive} was a special case of their method for uncorrelated data.
To account for heterogeneous data distributions, \cite{Ye2021} constructed a nonparametric CUSUM monitoring scheme and applied Thompson sampling for adaptive sampling. However, all of these works do not consider data autocorrelation. Consequently, they may ignore the autocorrelation information and delay the alarm. So far to our best knowledge, the only work considering temporal characteristics of the data for POMDSCD is the spatio-temporal monitoring of \cite{gomez2021adaptive}. It divides the streaming data into small overlapped windows, and for each window uses recursive tensor smooth-sparse completion for spatio-temporal structure learning and missing data estimation. The estimated sparse part is regarded as an abnormal change to design monitoring and adaptive sampling policies. This formulation does not describe temporal autocorrelation using a statistical model, but uses window-sliced tensor decomposition. 
Though it is flexible to deal with high-dimensional nonstructural temporal data,  it cannot track data autocorrelation parametrically or describe the evolution of the change, and consequently may lose certain detection power.  

\subsection{Multi-armed bandit (MAB)} 
For an  MAB problem, the bandit incrementally collects data for multiple arms (variables) in a decision space, and adaptively allocates sensing resources among multiple arms. There is an unknown correspondence between the reward and the arms that we select each time, but the more times we select an arm, the more information about its reward we know. For each selection, we can choose some arms that we have selected many times to get a stable reward or we can also choose some arms that we have selected few times to get more information about their reward to help us make the next decision. The goal of the bandit is to maximize the total rewards from the selected arms in a certain time period. Many works for MAB have been proposed for the best arm or top-K arm identification \cite{even2006action,bubeck2013multiple}, since the best arms also indicate best rewards. Among these methods, UCB is most commonly used for conducting a sampling strategy. It constructs a confidence interval for each arm based on its past observations, i.e., doing exploration. Then the arms with the highest upper confidence bounds are selected for the next observation, i.e., doing exploitation. There are also extensions of UCB to consider further adding adaptiveness to the confidence level. For example, \cite{guo2019adalinucb} and \cite{wu2018adaptive} provided to adaptively balance the importance between exploitation and exploration.

The problem of autocorrelation of rewards in different time points is also considered, as restless MAB (RMAB \cite{guha2010approximation,gafni2020learning}), where the state (i.e., expectation of the reward) of each arm continues evolving over time even when it is not played. Models such as linear dynamic models \cite{liu2012learning}, Markovian process \cite{dance2019optimal}, etc, have been applied to describe the evolving process of arms. Compared with traditional MAB, the policy of RMAB usually suggests conducting exploitation of an arm for certain several continuous time points and then updating its distribution inference, and decide which arm to exploit next according to the inference results.  

For the change point problem, there is also a branch of works for it in MAB, i.e., assuming rewards of different arms change piece-wisely \cite{liu2018change,cao2019nearly}, and constructing sampling policies that can track changes in time such that the best arms can always be selected. These methods can be potentially used in our problem to narrow down the attention to fewer variables which have more serious changed patterns. However, they can only help isolate the changed variables, i.e., do diagnosis, but cannot guarantee a quick system-level change point detection, since their sampling allocation strategies mainly focus on accurate arm estimation, instead of quick system-level change point detection. 

\section{Problem Formulation}
\label{sec:formulation}
With the superiority and easy interpretability of state space models in modeling autocorrelated data in engineering applications, we use state space model for multivariate data stream $\Y_{t}\in \calR^{p}, t=1,\ldots$. Specifically, we introduce a latent state variable $\X_{t}\in \calR^{q}$ and have
\begin{align}
\label{eq:space}
\Y_{t} & = \C\X_{t} + \bv_{t}, \\
\label{eq:state}
\X_{t} & = \A \X_{t-1} + \w_{t}. 
\end{align} 
Here $\A\in \mathcal{R}^{q \times q}$ is the state transition matrix which expresses the spatiotemporal autocorrelation between the hidden variables and $\C \in \mathcal{R}^{p \times q}$ is the output matrix which expresses the correlation between the observed and hidden variables at the same time point. $\bv_{t}\in \mathcal{R}^{p} \sim N_{p}(\bf{0},\bR)$, and $\w_{t}\in \mathcal{R}^{q}\sim N_{q}(\bf{0},\bQ)$. Here $\bv_{t}$ and $\w_{t}$ denote the observation noise and the state transfer noise, which are independent with each other and also with $t$. $\bQ=\sigma_{q}^{2}\bI$ and $\bR=\sigma_{r}^{2}\bI$ are diagonal matrices.  $p$ can be either larger or smaller than $q$. If $q=p$ and both $\A$ and $\C$ are diagonal matrices, the $p$ variables degenerate to be uncorrelated.

Due to limited sensing resources, at each sensing time $t$, we assume that we can observe only $m \ (m \leq p)$ out of the $p$ variables. By introducing the binary decision variable $z_{it}$ for each variable $Y_{it}$ such that $z_{it}=1$ if and only if $Y_{it}$ is observed at time $t$, the sensing constraint can be expressed as $\sum_{i=1}^{p}z_{it}=m, \forall t$. We further denote $Z(t)=\{\forall i, z_{it}=1\}$ as the vector of indices corresponding to the observed dimensions for $\Y_{t}$. 

We can use historical data to obtain parameter estimates for the state space model by the EM algorithm \cite{smith2003estimating, mader2014numerically}, which is a well-established estimation algorithm that has been studied, so in the subsequent sections we consider $\mathbf{A}$ and $\mathbf{C}$ to be known.

We would like to detect any change in the $p$ variables. Without loss of generality, we assume that a change $\fb \in \calR^{q}$ can occur on $\X_{t}$ since an unknown change point $\tau$, i.e., 
\begin{align}
\X_{t} &= \A\X_{t - 1}+\w_{t},       t < \tau, \\
\X_{t} &= \A\X_{t - 1}+\fb+\w_{t}, t \geq \tau.
\end{align}
$\fb$ is a general vector and we do not specify a pattern, and our goal is to construct a detection scheme to quickly detect $\tau$ based on sequential partial observations of $\Y_{t}, t= 1,\ldots$. 
We call the system is under in-control (IC) condition when $t < \tau$ and out-of-control (OC) condition when $t \geq \tau$.
\section{Methodology}
\label{sec:methodology}
In this section, we first propose the Kalman filtering algorithm for partially observed $\Y_{t}$ in Section \ref{sec:partialKF}. Then we propose the monitoring statistics based on GLRT in a change point detection framework in Section \ref{sec:change_point}. Then we propose an adaptive sampling policy based on UCB with the balance of exploration and exploitation in Section \ref{sec:UCB}. Finally, some properties of the adaptive sampling policy are discussed in Section \ref{sec:theory}. 
\subsection{State inference based on partially observable Kalman filter}
\label{sec:partialKF}
Under IC condition, for each time point $t$,  we can denote the observed dimensions as $\Y_{Z(t)}$   for time $t$, and unobserved dimensions as $\Y_{Z^c(t)}$ for time $t$. We can reformulate Equation (\ref{eq:space}) and (\ref{eq:state}) as follows
 \begin{align}
\begin{pmatrix} 
\Y_{Z(t)} \\ \Y_{Z^c(t)}
\end{pmatrix}
=
\begin{pmatrix} 
\C_{Z(t)} \\ \C_{Z^c(t)}
\end{pmatrix}
\X_{t} + 
\begin{pmatrix} 
\bv_{Z(t)} \\ \bv_{Z^c(t)}
\end{pmatrix},
 \end{align}
where $\C_{Z(t)} \in \calR^{m\times q}$ and $\C_{Z^c(t)} \in \calR^{(p-m)\times q}$. Based on $\Y_{Z(t)} , t= 1,\ldots, n$, we aim to infer $\X_{t},  t = 1, \ldots, n$ and its change accordingly, which can be achieved via the following proposed partially-observable Kalman filter. In particular, denote $\hat{\X}_{t+1|t}=\E(\X_{t+1}|\Y_{Z(1)},\ldots,\Y_{Z(t)})$ ,  we have the following recursive estimation  
\begin{align}
\label{for:kalman_1} 
\hat{\X}_{t+1|t} &= \tilde{\A}_{t} \hat{\X}_{t|t-1} + \A \bK_{t}\Y_{Z(t)},  \\
\label{for:kalman_2}
\tilde{\A}_{t} &= \A(\bI-\bK_{t}\C_{Z(t)}), \\
\label{for:kalman_3}
\bK_{t} &= \bP_{t|t-1}\C_{Z(t)}^{\prime}(\C_{Z(t)}\bP_{t|t-1}\C_{Z(t)}^{\prime}+\bR)^{-1}, \\
\label{for:kalman_4}
\bP_{t|t-1} &= \tilde{\A}_{t-1}\bP_{t-1|t-2} \tilde{\A}_{t-1}^{\prime} + \bQ.
\end{align}

Similarly with the classic Kalman filter, after observing partial dimensions of $\Y_t$ each time, we can calculate the Kalman gain coefficient $\bK_t$ using Equation (\ref{for:kalman_3}). However, it should be noted that the $\C_{Z(t)}$ in different time $t$  consist of different $m$ rows of the output matrix $\C$.  Then we can use Equation (\ref{for:kalman_2}) to calculate the updated estimated transition matrix $\tilde{\A}_{t}$.  After getting all of the above, we can predict the next $\X_{t+1}$ and the next covariance matrix $\bP_{t+1|t}$ using Equation (\ref{for:kalman_1}) and (\ref{for:kalman_4}).  This is also similar to the classic Kalman filter, except that $\Y_{Z(t)}$ has only $m$ dimensions of $p$. We need to note that the partial observation $\Y_{Z(t)}$ leads to less information compared with the classic Kalman filter, which may decrease the accuracy of estimation. This is a challenge for the partially-observable Kalman filter. Hence in the OC condition, we need to choose the best $m$ dimensions to observe and get more information about the anomaly. We will introduce an adaptive sampling strategy to solve the problem later. Last, we can use the above formulations for predicting $\hat{\Y}_{Z(t+1)}=\E(\Y_{Z(t+1)}|\Y_{Z(1)},\cdots,\Y_{Z(t)})$ as 
\begin{align}
\label{eq:residual}
\hat{\Y}_{Z(t+1)} &= \C_{Z(t+1)}\hat{\X}_{t+1|t}.
\end{align}

\subsection{Change point detection}
\label{sec:change_point}
Based on Equation \eqref{eq:residual},
we can consequently get the prediction error $\br_{t} = \Y_{Z(t)} - \hat{\Y}_{Z(t)}$.  It can reflect the untraceable signal of $\Y_t$, which includes system inherent fluctuations and change information that cannot be described by the IC model. As such, we can use the prediction error to generate monitoring statistics \cite{apley1999glrt, pan2004applying}. In particular, under IC condition, we can derive
\begin{align}
\label{eq:H0}
\E(\br_{t}) &= \mathbf{0}, 
\bV_{t} \equiv \Cov(\br_{t}) = \C_{Z(t)}\bP_{t|t-1}\C_{Z(t)}^{\prime} + \bR.
\end{align}
Under OC condition, we have
\begin{align}
\label{eq:H1}
\E(\br_{t}) = \C_{Z(t)}\tilde{\bg}_{t}, \bV_{t} = \C_{Z(t)}\bP_{t|t-1}\C_{Z(t)}^{\prime} + \bR,
\end{align}
where $\tilde{\bg}_{t} = \tilde{\A}_{t} \tilde{\bg}_{t-1} + \fb$. Clearly, if there occurs a shift $\fb$, $\tilde{\bg}_{t}$ will accumulate over time, which leads to a larger $\br_{t}$. Consequently, we can use this to build the monitoring statistics. 

In particular, suppose the current sensing time is $n$, we have the past data $\Y_{Z(t)}, t= 1,\ldots,n$, and would like to detect whether there has been a change before $n$, i.e., 
\begin{align}
H_{0}: \tau \geq n , H_{1}: 0<\tau < n.
\end{align}
Suppose there is an assumed change point $0<k<n$. With the derived IC and OC distribution of $\br_t$, we can derive the likelihood ratio test statistic for the hypothesis as
\begin{equation}
\begin{aligned}
l(n,k,\fb) &= \log \frac{\prod_{t=1}^{k}f_{0}(\br_{t})\cdot \prod_{t=k+1}^{n}f_{1}(\br_{t})}{\prod_{t=1}^{n}f_{0}(\br_{t})}\\
&= \sum_{t=k+1}^{n}\log \frac{f_{1}(\br_{t})}{f_{0}(\br_{t})},
\end{aligned}
\end{equation}
where $f_{0}(\br_{t})$ denotes the probability of $\br_{t}$ under $H_{0}$ in Equation (\ref{eq:H0}), and $f_{1}(\br_{t})$ denotes the probability of $\br_{t}$ under $H_{1}$ in Equation (\ref{eq:H1}). 
In our case, we have 
\begin{equation}
\begin{aligned}
\label{eq:likelihood}
\ell(n,k,\fb)  &= \frac{1}{2}\sum_{t=k+1}^{n}\br_{t}^{\prime}\bV_{t}^{-1}\br_{t}\\ 
&-\frac{1}{2}\sum_{t=k+1}^{n}(\br_{t}-\C_{Z(t)}\tilde{\bg}_{t,k})^{\prime}\bV_{t}^{-1}(\br_{t}-\C_{Z(t)}\tilde{\bg}_{t,k}).
\end{aligned}
\end{equation}
Consider that $\tilde{\bg}_{t,k}$ is iteratively defined, we can reformulate its expression as: 
\begin{align}
\tilde{\bg}_{t,k}  = (\sum_{s=k+1}^{t}\prod_{l=k}^{s-1}\tilde{\A}_{l}+\bI)\fb = \bG(t,k)\fb.
\end{align}
Bringing it back to Equation (\ref{eq:likelihood}), we get
\begin{equation}
\begin{aligned}
\label{eq:lrt}
\ell(n,k,\fb) &\propto \sum_{t=k+1}^{n}(\C_{Z(t)}\bG(t,k)\fb)^{\prime}\bV_{t}^{-1}\br_{t} \\
&+ \frac{1}{2}\sum_{t=k+1}^{n}\fb^{\prime}\bG(t,k)^{\prime}\C_{Z(t)}^{\prime}\bV_{t}^{-1}\C_{Z(t)}\bG(t,k)\fb.
\end{aligned}
\end{equation}
Since $\fb$ is unknown, we replace it with a maximum likelihood estimator for fixed values of $k$ and $n$ in Equation (\ref{eq:lrt}). Taking derivative of $\ell(n,k,\fb)$ with respect to $\fb$ and setting it to zero, we obtain the estimation
\begin{align}
\label{eq:f_hat}
\hat{\fb}(n,k) = \bSigma_{f}(n,k)(\sum_{t=k+1}^{n}\bG(t,k)^{\prime}\C_{Z(t)}^{\prime}\bV_{t}^{-1}\br_{t} ),
\end{align}
where $\bSigma_{f}(n,k)=(\sum_{t=k+1}^{n}\bG(t,k)^{\prime}\C_{Z(t)}^{\prime}\bV_{t}^{-1}\C_{Z(t)}\bG(t,k))^{-1}$. Then we have the GLRT statistic by replacing  $\hat{\fb}(n,k)$ in Equation (\ref{eq:likelihood}) as
\begin{equation}
\begin{aligned}
\label{for:statistic}
\ell(n,k, \hat{\fb}(n,k)) = &(\sum_{t=k+1}^{n}\bG(t,k)^{\prime}\C_{Z(t)}^{\prime}\bV_{t}^{-1}\br_{t} )^{\prime}\bSigma_{f}(n,k)\\
&(\sum_{t=k+1}^{n}\bG(t,k)^{\prime}\C_{Z(t)}^{\prime}\bV_{t}^{-1}\br_{t}  ).
\end{aligned}
\end{equation}

\begin{proposition}
\label{prop:IC} 
If the null hypothesis is true, the GLRT statistic $\ell(n,k, \hat{\fb}(n,k)) $ follows a $\chi^{2}$ distribution with $p$ degrees of freedom. 
\end{proposition}
\begin{proposition}
\label{prop:OC} 
If the alternative hypothesis is true and $k=\tau$ is the true change point, $\hat{\fb}(n,\tau)$ follows a Gaussian distribution with 
\begin{align}
\label{eq:mean_f}
\E(\hat{\fb}(n,\tau)) &=\fb, 
\Cov(\hat{\fb}(n,\tau)) = \bSigma_{f}(n,\tau).
\end{align}
Then $\ell(n,k, \hat{\fb}(n,\tau))$ follows a non-central chi-square distribution $\chi_{p}^{2}(c_{n}^{\tau})$ of $p$ degrees of freedom
with the non-centrality 
\begin{align}
\label{eq:non-central}
c_{n}^{\tau} &= \fb^{\prime} \bSigma_{f}(n,\tau)^{-1} \fb.
\end{align}
\end{proposition}
Proposition \ref{prop:OC} indicates large values of $\ell(n,k, \hat{\fb}(n,k))$ is evidence against the null hypothesis. Since the change point location $\tau$ is unknown, we can use a constrained MLE estimator, i.e., $\hat{\tau}=\arg\max_{n-m_{1}<k<n-m_{2}} \ell(n,k, \hat{\fb}(n,k))$ where $m_1$ and $m_2$ are pre-defined values. This indicates we consider at least the past $m_{2}$ samples and at most the past $m_{1}-1$ as potential change points from the current time $n$. The role of the window size is two-fold. On the one hand, it effectively ensures the stability of the MLE estimator. On the other hand, it reduces the memory requirements to implement the detection procedure in reality \cite{cao2019sketching,xiesequential}. 

Then we can define a scanning scheme with a stop time that triggers an OC alarm as soon as the GLRT statistic rises above a threshold $h>0$: 
\begin{align}
\label{eq:changepoint}
N = \inf\{n: \ \ell(n,\hat{\tau}, \hat{\fb}(n,\hat{\tau}))>h\},
\end{align}
with $N$ defined as the run length. Generally, we would like $N$ to be large enough when the process is under IC condition. In contrast, we hope $N$ to be small when the process has changed to OC. Of course, this is related to the adaptive sampling strategy, which is designed in the next section. 
\subsection{Adaptive sampling with upper confidence region}
\label{sec:UCB}
Suppose up to time point $n$, the OC alarm has not been triggered. We need to decide $Z(n+1)$. According to Proposition \ref{prop:IC} and Proposition \ref{prop:OC}, a larger non-centrality indicates a larger difference between normal and changed cases. As such, the optimal adaptive sampling scheme should consider maximizing Equation (\ref{eq:non-central}). 
If $\fb$ and $\tau$ are known, we can develop the sequential decision rule which can maximize 
\begin{align}
\label{eq:optimal_Z}
Z_{n+1} = \arg \max_{Z} \fb^{\prime} (\bG(n+1,\tau)^{\prime}\C_{Z}^{\prime}\bV_{n+1}^{-1}\C_{Z}\bG(n+1,\tau))\fb.
\end{align} 
Unfortunately, $\fb$ and $\tau$ are unknown.
An alternative is to replace $\fb$ and $\tau$ in Equation (\ref{eq:optimal_Z}) by its estimation $\hat{\fb}(n,\hat{\tau})$ and $\hat{\tau}$. 
Yet we need to consider the influence of its estimation uncertainty on Equation (\ref{eq:optimal_Z}) due to the limited observations we have. Enlightened by UCB, we propose the following upper confidence region algorithm. In particular, consider the $1-\alpha$ confidence region of $\hat{\fb}(n,\hat{\tau})$ as the ellipse corresponding to 
\begin{equation*}
\begin{aligned}
P[&\fb \text{ fall inside } \{(\fb-\hat{\fb}(n,\hat{\tau}))^{\prime}\bSigma_{f}(n,\hat{\tau})^{-1}(\fb-\hat{\fb}(n,\hat{\tau})) 
\\&= \chi_{1-\alpha}^{2}(q)\}]=1-\alpha.
\end{aligned}
\end{equation*}
Define $\bOmega_{Z}(n+1,\hat{\tau})=\bG(n+1,\hat{\tau})^{\prime}\C_{Z}^{\prime}\bV_{n+1}^{-1}\C_{Z}\bG(n+1,\hat{\tau})$ and $\bV_{n+1} = \C_{Z}\bP_{n+1|n}\C_{Z}^{\prime} + \bR$. 
Our goal is to find the optimal $Z_{n+1}$ which can maximize the following criterion: 
\begin{align}
\label{eq:strategy_Z}
& Z_{n+1}   = \arg \max_{Z,\fb} \fb^{\prime} \bOmega_{Z}(n+1,\hat{\tau})\fb, \\
\label{eq:strategy_Z_constraint}
\text{s.t. } & (\fb-\hat{\fb}(n,\hat{\tau}))^{\prime}\bSigma_{f}(n,\hat{\tau})^{-1}(\fb-\hat{\fb}(n,\hat{\tau})) = \chi_{1-\alpha}^{2}(q).
\end{align} 
The objective function aims to exploit the most changed dimensions based on the current estimation of $\hat{\fb}(n,\hat{\tau})$ for exploitation, while the constraints consider the uncertainty of $\hat{\fb}(n,\hat{\tau})$ for exploration. 
To be more specific, similar to the UCB in the one-dimensional case, since the estimate $\hat{\fb}(n, \hat{\tau})$ has variance and deviates from the true value, we search for the optimal solution on the boundary of the high-dimensional ellipsoid expressed by Equation (\ref{eq:strategy_Z_constraint}). In this way, we can make use of the estimate while considering the inaccuracy due to its variance.
The above optimization problem can be solved in a two-step approach. First, given a particular $Z$, find the  $\fb$ that can maximize Equation (\ref{eq:strategy_Z}) under the constraint of Equation (\ref{eq:strategy_Z_constraint}), which can be solved efficiently based on Proposition \ref{prop:find_f}. Second, enumerate all possible $Z\in \calZ$ where $\mathcal{Z}$ is the set of all the $\tbinom{p}{m}$
combinations of the choice of $m$ from $p$ and select the best one.

\begin{proposition}
\label{prop:find_f}
We can perform a Cholskey decomposition of $\bSigma_{f}(n,\hat{\tau})$,
\begin{align}
    \bSigma_{f}(n,\hat{\tau}) = \mathbf{B} \mathbf{B}^T,
\end{align}
where $\mathbf{B}$ is orthogonal and $\mathbf{B}^T\bSigma_{f}(n,\hat{\tau})^{-1}\mathbf{B} = \mathbf{I}$. Then for  a fixed $Z\in \calZ$, we can obtain the following eigenvalue decomposition,
\begin{align}
    \mathbf{B}^T\bOmega_{Z}(n+1,\hat{\tau})\mathbf{B} = \mathbf{D}^T \bLambda \mathbf{D}.
\end{align}
Then we can define $\mathbf{H} = \mathbf{B}\mathbf{D}^T$ and get the following equations
\begin{align}
\mathbf{H}^T \bSigma_{f}(n,\hat{\tau})^{-1} \mathbf{H} &= \mathbf{I},\\
\mathbf{H}^T \bOmega_{Z}(n+1,\hat{\tau})\mathbf{H} &= \bLambda,
\end{align}
where $\bLambda$ is a diagonal matrix and $\lambda_i$ is its $i$th diagonal element. $\lambda$ is the solution of $\sum_{i = 1}^{p}\frac{x_{i}^2}{(\lambda_{i} + \lambda)^2} = \chi_{1-\alpha}^{2}(q)$, where $x_i$ is the $i$th element of $\x = \bLambda \bH^{-1} \hat{\fb}(n,\hat{\tau})$. Finally, we can get $\tilde{\fb}$ via solving the following
\begin{align}
    \Vert \tilde{\fb} \Vert_{2}^{2} &= \chi_{1 - \alpha}^{2}(q),\\
    (\bLambda + \lambda \I)\tilde{\fb} &= -\bLambda\bH^{-1}\hat{\fb}(n,\hat{\tau}).
\end{align}
Then we can get $\fb = \mathbf{H}\tilde{\fb} + \hat{\fb}(n,\hat{\tau})$ as the solution of Equation (\ref{eq:strategy_Z}) under constraint of Equation (\ref{eq:strategy_Z_constraint}).
\end{proposition}

We need to note that $\alpha$ in Equation ({\ref{eq:strategy_Z}}) determines the size of the confidence region. For a small $\alpha$, the region will be large and the algorithm will focus more on exploration. For a large $\alpha$, the region will be small and the algorithm will focus more on exploitation. Generally, $\alpha$ can be determined as a tuning parameter in advance. Yet another better alternative is to set $\alpha$ adaptively. As we know, for change point detection purpose, when the test statistic $T_{n}= \ell(n,\hat{\tau}, \hat{\fb}(n,\hat{\tau}))$ in Equation (\ref{eq:changepoint}) is small, we hope to explore more since we have large confidence that the system is normal and the change point occurred in the system with small chance.  In contrast, when $T_{n}$ is large, we need to exploit more since we have large confidence that the system already has a change point. Hence an adaptive $\alpha$ which is associated with $T_{n}$ is more reasonable to balance between exploration and exploitation. As such, we set an adaptive $\alpha_{n}$ with the form of
\begin{align}
    \label{eq:alpha setting_linear}
    \alpha_{n} = \min(\max(\frac{T_{n} - d}{l}, 0) + \alpha_{min}, \alpha_{max}),
\end{align}
which is a segment-wise linear function specified by $\left\{d, l,\alpha_{min}, \alpha_{max}\right\}$. $l$ is a scale parameter which determines the rate of change of $\alpha_n$, $d$ determines how much $T_n$ is reached when $\alpha_n$ starts to change and $\alpha_{max}$, $\alpha_{min}$ represent the upper and lower bounds of the change in $\alpha_n$.

Hereafter, we denote our proposed detection scheme based on Adaptive Upper Confidence Region with State Space model as AUCRSS. Its adaptive sampling procedure is shown in Algorithm \ref{alg:prompt_sampling}. The whole procedure of AUCRSS is shown in Algorithm \ref{alg:total_alg}. 

\begin{algorithm}
		\caption{Adaptive sampling based on AUCRSS}
        \begin{algorithmic}
		 \REQUIRE $\fb(n,\hat{\tau})$, $\bSigma_{\fb}(n,\hat{\tau})$, $\bV_{n}$,$\bG(n,\hat{\tau})$, estimated upon to the current time point $n$
		 \ENSURE $Z(n+1)$
		 \FORALL {$Z_{k}\in \calZ$}
		  \STATE Calculate $\bOmega_{Z_k}(n+1)$
		 and find the best $\tilde{\fb}$ according to Proposition \ref{prop:find_f} \\
		  \STATE Get  $Score_k = \tilde{\fb}^{\prime} \bOmega_{Z_k}(n+1, \hat{\tau})\tilde{\fb}$
		 \ENDFOR
	  \STATE select $Z_k$ with the largest $Score_k$ as $Z(n + 1)$
    \end{algorithmic}
    \label{alg:prompt_sampling}
	\end{algorithm}

\begin{remark}
For process with large $p$ and $m$, enumerating all the possible $\bOmega_{z}(n+1,\hat{\tau})$ may be time consuming. An empirical solution is to select the $m$ dimensions one by one.  We initially set $Z$ to be empty, with its complementary set as $Z^{c}$. We use $m$ iterations to add dimensions into $Z$ sequentially. In each iteration, for each $k \in Z^{c}$, we set $Z_{k}=\{k\cup Z\}$, and calculate its $\bOmega_{Z_{k}}(n+1,\hat{\tau})$. Then we   calculate its best $\tilde{\fb}$ according to Proposition \ref{prop:find_f} and get $Score_k = \tilde{\fb}^{\prime} \bOmega_{Z_k}(n+1, \hat{\tau})\tilde{\fb}$. Finally we select  $k^{\star} = \arg\max_{k\in Z^{c}}Score_{k}$, and update $Z \leftarrow \{Z\cup k^\star\}$ and $Z^{c} \leftarrow Z^{c}\setminus k^\star$. We name the simplified algorithm as Empirical AUCRSS, denoted as E-AUCRSS. Its detailed sampling algorithm is shown in Algorithm
\ref{alg:e-prompt_sampling}. 
\end{remark}

\begin{remark}
In Algorithm \ref{alg:total_alg}, we need to collect information about the system first before we start monitoring. So we assume that there will be no change in the first $n_0$ time points of the system, and we can perform random sampling about the dimensions for these $n_0$ time points for collecting information.
\end{remark}

	\begin{algorithm}
		\caption{ Adaptive sampling based on E-AUCRSS}
  \begin{algorithmic}
		\REQUIRE $\fb(n,\hat{\tau})$, $\bSigma_{\fb}(n,\hat{\tau})$,$\bV_{n}$,$\bG(n,\hat{\tau})$, estimated upon to the current time point $n$
		\ENSURE $Z(n+1)$
		 \STATE Set $Z = \emptyset$, and $Z^c = \{1,\ldots,p\}$
		\WHILE {$|Z|<m$}
		    \FOR{$k \in Z^{c}$}
		    \STATE Set $Z_{k}=\{k \cup Z\}$ and get $\bOmega_{Z_{k}}(n+1,\hat{\tau})$\\
		  \STATE  Find its best $\tilde{\fb}$ according to Proposition \ref{prop:find_f} \\
	 \STATE  Get  $Score_k = \tilde{\fb}^{\prime} \bOmega_{Z_k}(n+1)\tilde{\fb}$
	    \ENDFOR
        \STATE Get $k^{\star} = \arg\max_{k\in Z^{c}}Score_{k}$\\
    	\STATE Set $Z \leftarrow \{Z\cup k^\star\}$, and $Z^{c} \leftarrow Z^{c}\setminus k^\star$\\
		\ENDWHILE
         \STATE Set $Z(n+1) = Z$

\end{algorithmic}
\label{alg:e-prompt_sampling}
\end{algorithm}

	\begin{algorithm}
		\caption{Change point detection scheme based on AUCRSS or E-AUCRSS}
    \begin{algorithmic}
		\REQUIRE Data streams $\textbf{Y}_{n}, n = 1,\cdots$
	    \FOR{$n = 1, \cdots, n_{0}$}
	    \STATE Set the initial sampling set $Z(n)$ by randomly selecting $m$ variables out of $p$ variables
	    \ENDFOR
	    \FOR{$n = n_{0} + 1, \cdots$}
	    \STATE Collect the data $\textbf{Y}_{Z(n)}$\\
	    \STATE Update the Kalman filter via Equations (\ref{for:kalman_1}), (\ref{for:kalman_2}), (\ref{for:kalman_3}) and(\ref{for:kalman_4})\\
	    \STATE Calculate the detection statistic $\ell(n,\hat{\tau}, \hat{\fb}(n,\hat{\tau}))$ via Equations (\ref{for:statistic}) and (\ref{eq:mean_f})\\
	    \IF{$\ell(n,\hat{\tau}, \hat{\fb}(n,\hat{\tau})) \geq h $}
	     \STATE Trigger an alarm\\
	    \ELSE
	    \STATE Decide the next sampling set $Z(n + 1)$ via Algorithm \ref{alg:prompt_sampling} or Algorithm \ref{alg:e-prompt_sampling}\\
	    \ENDIF
	    \ENDFOR
     \end{algorithmic}
     \label{alg:total_alg}
\end{algorithm}

\subsection{Some properties of AUCRSS}
\label{sec:theory}
Now we analyze some asymptotic properties of AUCRSS under certain conditions. Because the state space evolving process is very complex, we only consider some specific cases. First, under IC conditions, we have the following results.
 
 \begin{theorem}
 \label{the:IC}
 Under the condition that $p = q$ and $\A$, $\C$ are diagonal matrices, when the process is IC, as the time point $n \to \infty$, suppose there exists an unbalanced dimension $k$ that has been only observed a finite number of times with order $O(1)$, then we have 
 \begin{align}
\label{Theorem1}
P(k \notin Z(n)) = O(\frac{1}{\sqrt{n}})
\end{align} 
The detailed derivation is shown in the Appendix. 
 \end{theorem}
 
Theorem \ref{the:IC} guarantees that if there exists some dimensions with an unbalanced and insufficient number of observations, our algorithm will prefer to observe these dimensions with a higher probability, thus ensuring that the sampling tends to be more balanced overall dimensions under IC condition.

Under OC conditions, suppose the real change point is $\tau$ and the true changed dimension set is $S^{**}$ with cardinality $|S^{**}|=s$, we define the optimal chosen set $Z^{**}$ with cardinality $|Z^{**}|=m$ as follows: if $s=m$, then $Z^{**}=S^{**}$; if $s \geq m$, $Z^{**}$ is the subset of $S^{**}$ composed of dimensions which can maximize $\fb^{T} \bOmega_{Z} \fb$; if $s \leq m$,  $Z^{**}$ is set to include $S^{**}$ and some other elements, which in this work are the fixed $m-s$ dimensions with the smallest variable indices in the complementary set of $S^{**}$. Suppose the following two assumptions are satisfied.

\begin{assumption}
There exists a matrix   $\bOmega_{max}$ that satisfies that $ \fb^{T}\bOmega_{max}\fb = sup\{\fb^{T}\{\bOmega_{Z^{**}}(n, \hat{\tau})\fb | _{n \geq 1} \}$. 
\end{assumption}

\begin{assumption}
There exists at least a good observation set $Z^{*}$ such that  $\fb^{T}\bOmega_{Z^{*}}(n, \hat{\tau})\fb \geq (1 - \Delta_{min})\fb^{T}\bOmega_{max}\fb$, with a certain small constant $\Delta_{min}>0$. 
\end{assumption}
We have the following two OC properties for AUCRSS.   
\begin{lemma}
\label{lemma_pt}
Under Assumptions 1-2, suppose $\A$ is a diagonal matrix, and $\C$ has exactly one non-zero element in each row and at least one non-zero element in each column. When the process has a change $\fb$ since $\tau=1$, denote $p_{n}$ as the probability of choosing $Z^{\star}$ for the time point $n$. When $\alpha \to 1$, we have $\lim_{n \to \infty} p_{n} = 1$, and 
\begin{align}
\label{eq:lemma1}
p_{n+1} \geq \frac{1}{\frac{1}{p_{0}} - \sum_{t = 0}^{n}\frac{1}{p_{t} + \frac{1}{2sa_{0}\tilde{\delta}}e^{\frac{1}{2}(a_{0}\tilde{\delta}(n_0+t))^{2}}}},
\end{align}
where $p_0$ is the probability to observe $Z^{*}$ after the random sampling of the first $n_0$ time. $a_{0}$ is the upper $(1 - p_{0}/2)$-quantile of the standard normal distribution and $\tilde{\delta}= (1 - \Delta_{min})\tilde{\sigma}_{max}$ where $\tilde{\sigma}_{max}=\max diag(\bOmega_{max})$ is the maximum diagonal element of $\bOmega_{max}$.
\end{lemma}
 The detailed derivation of Lemma \ref{lemma_pt} is shown in the Appendix. 

Lemma \ref{lemma_pt} provides the lower bound of $p_{n}$. In the Appendix, we can show that $p_{n}$ is a monotone nondecreasing sequence. Since $p_{n}$ has an upper bound $1$,  $p_{n}$ must converge as $n$ approaches infinity. In reality, we can randomly sample variables in the first $n_0$ times until $p_{0} $ reaches $ \frac{1}{1 + s\sqrt{2\pi}}$. Consequently, $p_{t}$'s lower bound can converge to 1 as $n$ increases when $p_{0} = \frac{1}{1 + s\sqrt{2\pi}}$, according to Equation (\ref{eq:lemma1}). 
This is consistent with our intuition, if the system is abnormal, the more time points we observe, the more likely we are to find anomalies of the system.
This guarantees that $p_{n}$ can converge to 1. Furthermore, we can find that the lower bound will increase when the initial probability $p_{0}$ increases or the number of shifted dimensions $s$ increases. This is reasonable since the higher the probability of initial observation or the more changed dimensions in the system, the easier it is to observe changed dimensions in subsequent observations.

\begin{theorem}
\label{the:OCARL_CMAB}
Under Assumptions 1-2, suppose $\A$ is a diagonal matrix, and $\C$ has exactly one non-zero element in each row and at least one non-zero element in each column. When the process has a change $\fb$ since $\tau=0$,   if $p_{0} = \frac{n_0}{n_0 + s\sqrt{2\pi}}$, then we can say the difference between the theoretical optimal value and our estimate is of constant order of magnitude as follows:
    \begin{align}
        \fb^{T} \bSigma_{f}(n,\hat{\tau})^{-1} \fb  = \fb^{T}n \bOmega_{max}\fb + O(1).
    \end{align}
\end{theorem}
As mentioned in Equation (\ref{eq:non-central}), $\fb^{T}\bSigma_{f}(n,\hat{\tau})^{-1}\fb$ denotes the non-centrality we have observed from the beginning to time $n$ under OC condition.  $\fb^{T} n\bOmega_{max}\fb$ denotes the maximum non-centrality we can observe theoretically from beginning to time $n$.  The increase rates of both of them are  $O(n)$. Theorem \ref{the:OCARL_CMAB} proves that their difference is yet $O(1)$, independent of $n$. This indicates that the difference is bounded to a value determined at the beginning of the data stream, and is mainly caused by random sampling. As $n$ increases and more information about change is collected, AUCRSS  guarantees that there will be no detection power difference between the test statistic based on the selected dimensions of AUCRSS and that based on the theoretically optimal dimensions. Consequently, their difference will not increase linearly but keep as constant. Theorem \ref{the:OCARL_CMAB} guarantees that after a finite time period, AUCRSS is able to choose the good dimensions to observe and trigger alarm in time. The detailed derivation of Theorem $\ref{the:OCARL_CMAB}$ is shown in the Appendix.

\section{Numerical Study}
\label{sec:numeric study}
This section conducts synthetic data experiments for the performance evaluation of AUCRSS and E-AUCRSS. 
We use $ADD_{IC} = E(T|\tau = \infty)$ to denote the average detection delay in IC conditions and $ADD_{OC} = E(T-\tau|T\geq \tau, \tau < \infty)$ to denote the average detection delay in OC conditions. 
In particular, we control the $ADD_{IC}$ to a specific value, and compare the $ADD_{OC}$ of different methods to evaluate their performance. For convenience, we refer to $ADD_{IC}$ and $ADD_{OC}$ as ADD.
For better performance illustration, the standard deviation of ADD, i.e., SDD, is also presented. 

The following state-of-the-art methods for AUCRSS are also considered as benchmarks: (1)
TRAS \cite{liu2015adaptive}: the top-r adaptive sampling detection algorithm which is based on the conventional univariate CUSUM statistics and a constant imputation parater to deal with partial observation; 
(2) CMAB \cite{zhang2019partially}: the combinatorial multi-armed bandit based adaptive sampling algorithm which uses MAB problem to deal with the partial observations and considers the correlation between different dimensions ; 
(3) CMAB(s) \cite{zhang2019partially}: the simplified version of CMAB without considering the correlations of different dimensions;
(4) SASAM \cite{wang2018spatial}: the spatial-adaptive sampling and monitoring algorithm which is based on the CUSUM statistics and leverages the spatial information of data for change point detection; 
(5) NAS \cite{xian2018nonparametric}: the nonparametric anti-rank adaptive sampling algorithm which uses a constant imputation parameter for tackling partial observations; 
(6) R-SADA \cite{xian2021online}: the rank-based sampling algorithm by data augmentation which is based on CUSUM statistics and uses a rank-based algorithm to deal with partial observation; 
(7) TSS \cite{gomez2021adaptive}: the tensor sequential sampling algorithm which uses an exponentially weighted moving average (EWMA) control chart to monitor the high-dimensional data and uses tensor recovery to deal with the partial observation; 
(8) R-AUCRSS: it has the same detection framework but randomly chooses $m$ dimensions to observe for each time point.
R-AUCRSS can be regarded as the lower performance bound of AUCRSS. Comparing it with other baselines can evaluate the importance of considering autocorrelation in streaming data. Comparing it with AUCRSS can evaluate the importance of considering an adaptive sampling strategy. 

We conduct our experiments for performance evaluation of different algorithms from the following three aspects: 1) the difference between an adaptive $\alpha_n$ and a constant $\alpha$ in AUCRSS; 2) the differences between AUCRSS, E-AUCRSS and  R-AUCRSS; 3) the difference between E-AUCRSS and other benchmarks.

\subsection{Influence of $\alpha_{n}$}
\label{sec:alpha}
In this subsection, we discuss the influence of adaptive $\alpha_n$ on the detection performance. We consider a $p=10$ dimensional process with $q=7$ dimensional latent variable. The transition matrix $\A$ and output matrix $\C$ are shown in the appendix, due to space constraints.
The standard deviation of the two errors are set as $\sigma_q = \sigma_r = 0.1$. For the OC change pattern, we set the true change point $\tau=0$, and $\fb = [f, 0, 0, 0 ,0, 0, 0]^{T}$ with different shift magnitudes $f = 0.05,\  0.06,\ ...,\  0.1$. 

\begin{figure*}[htb]
    \centering
    \subfloat[]{%
   
        \includegraphics[scale = 0.061]{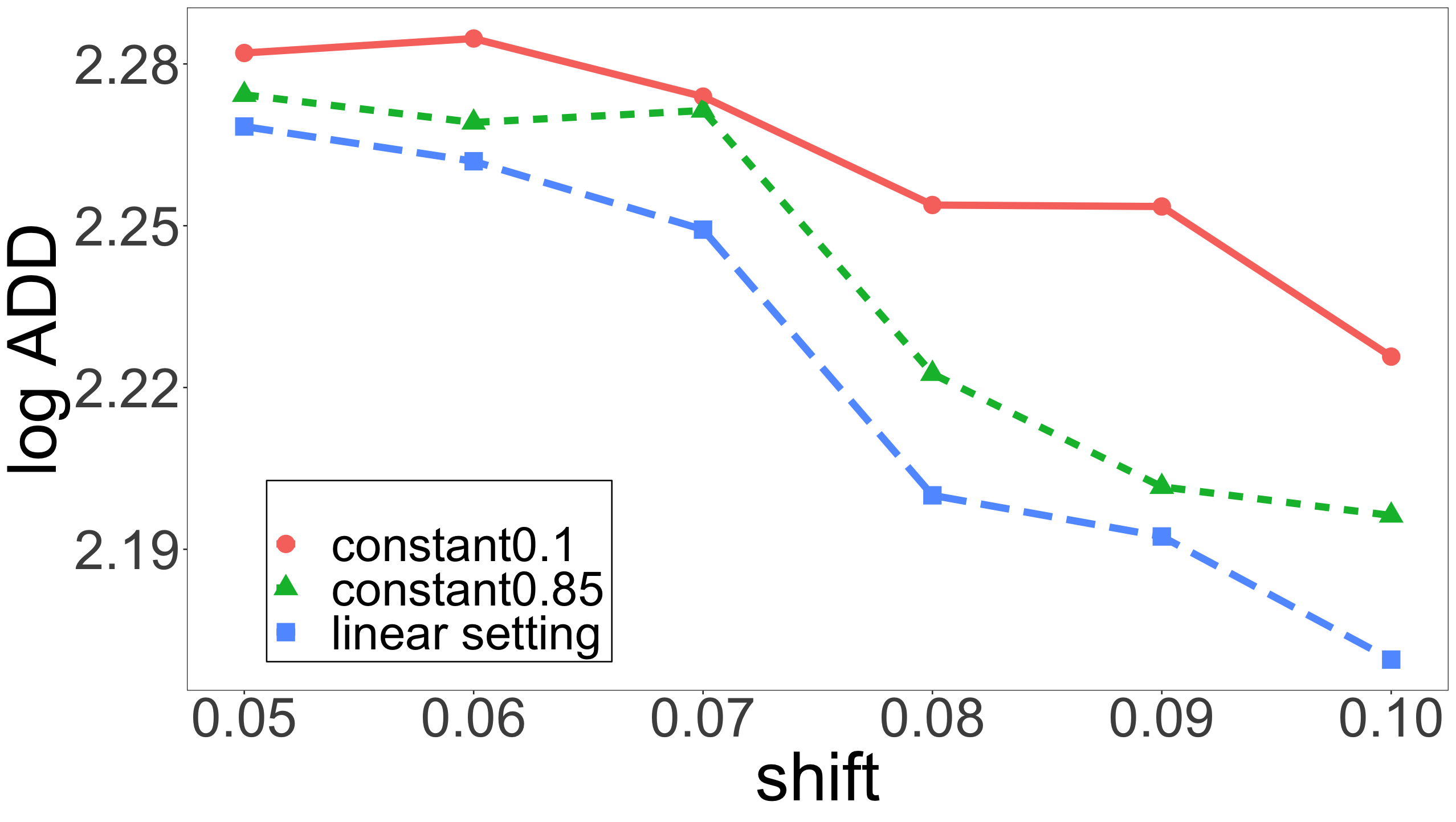}
        \label{fig:linear_bestm_1}
    }
    \subfloat[]{
 
        \includegraphics[scale = 0.061]{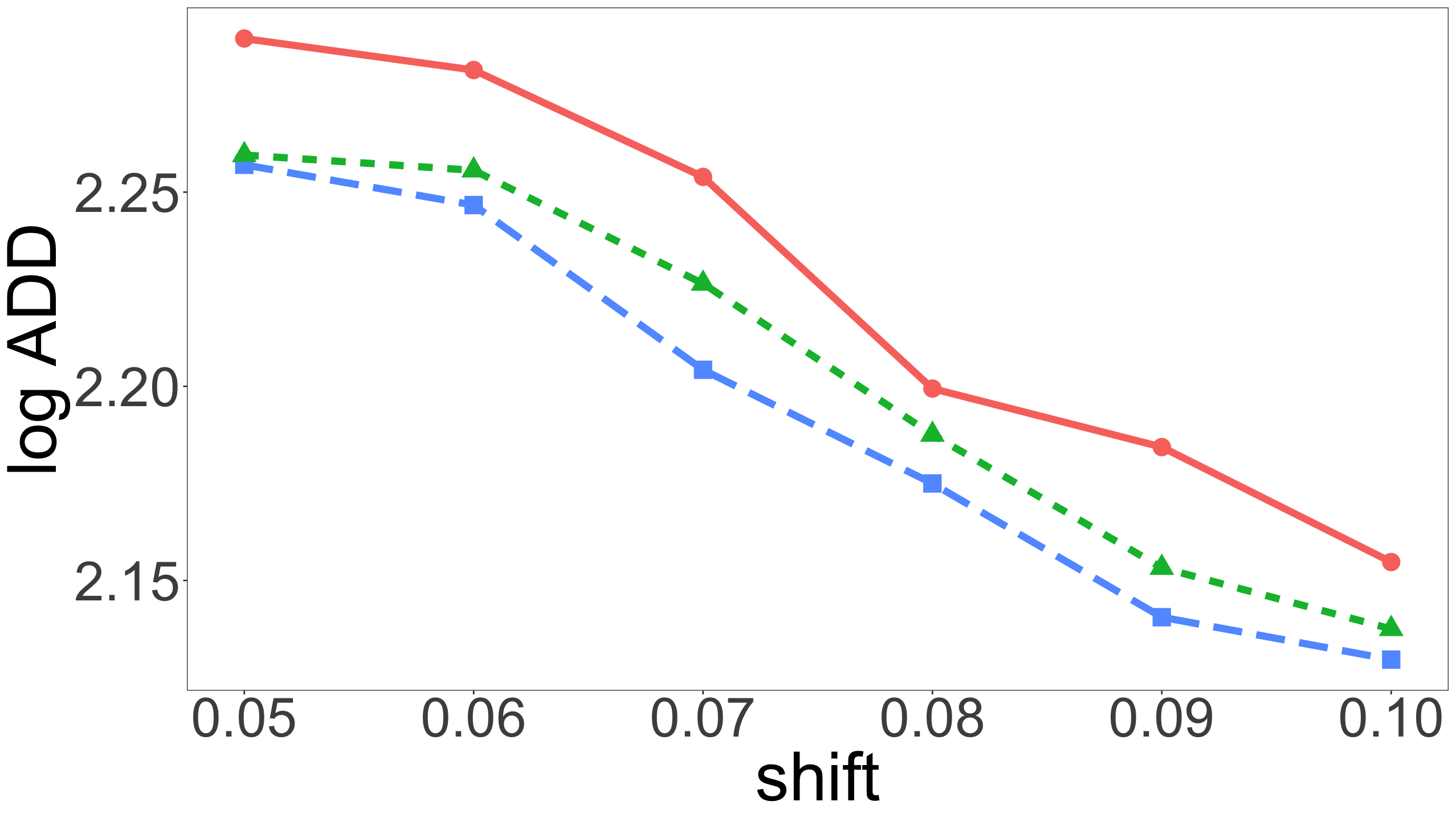}
        \label{fig:linear_bestm_2}
    }
    \subfloat[]{%
   
        \includegraphics[scale = 0.061]{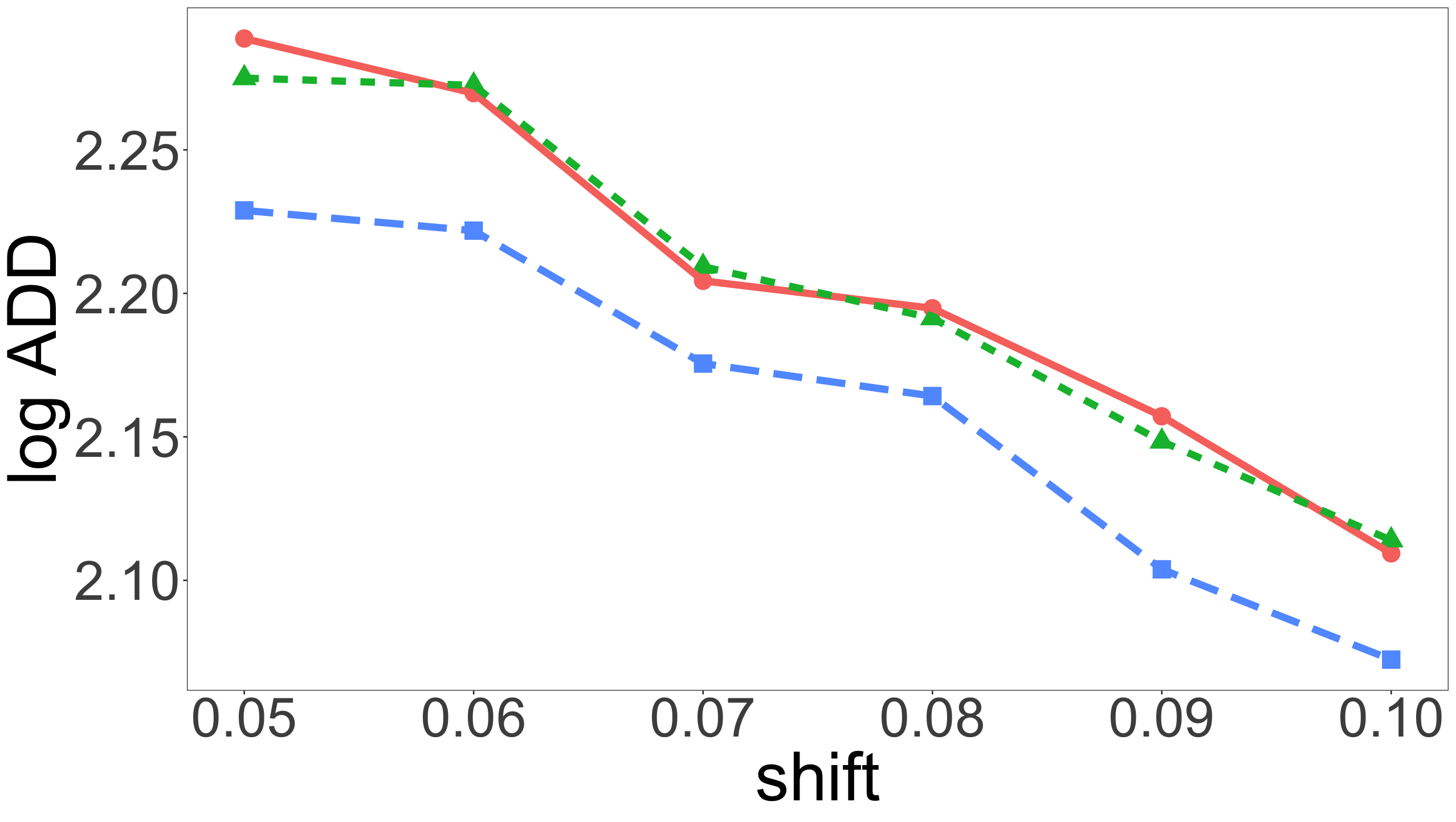}
        \label{fig:linear_bestm_3}
    }
    \caption{Comparison between segment-wise linear $\alpha_n$ and constant $\alpha_n$, for $10$-dimensional data streams with (a) $m = 1$, (b) $m = 2$ and (c) $m = 3$.}
    \label{fig:linear_best}
\end{figure*}

For AUCRSS-based methods, we set the time windows $m_1 = 50$ and $m_2 = 5$, and consider three settings of $m = 1, 2, 3$ respectively. We consider an adaptive segment-wise linear $\alpha_n = \min(\max(\frac{T_{n} - 15}{6.67}, 0) + 0.1, 0.85)$ and another two constant $\alpha_n = 0.1$ and $\alpha_n = 0.85$. For each $\alpha_n$ under each $m$, the specific selection method of $h$ is as follows. We simulate the monitoring process for in total $N$ replications (say $N = 50,000$ used in this paper). Then we use the bisection algorithm \cite{Qiu2010nonparametric} to search the control limit $h$ so that the $ADD_{IC}$ of these N replications equals 200.
Controlling the value of $ADD_{IC}$, we can evaluate the OC performance of different $\alpha_n$ in Figure \ref{fig:linear_best}. We can find that the adaptive $\alpha_{n}$  has better performance than the constant $\alpha_n$ for all the cases. A too small constant $\alpha_{n} = 0.1$ leads to extra exploration and an underuse of information already gotten. In contrast, a too large constant $\alpha_{n} = 0.85$ leads to extra exploitation and underuse of new information. By using the linear setting of $\alpha_n$, we can avoid the situations above and make an adaptive balance between exploitation and exploration.

Besides the above segment-wise linear setting of $\alpha_n$, we also considered other settings of $l$, $d$, $\alpha_{min}$ and $\alpha_{max}$ in our experiments. Yet we find that the performance under different linear settings is similar without significant differences. This demonstrates that the performance is very robust to different linear settings for $\alpha_n$, and in practice, it is very flexible and convenient to set these parameters. In addition, it should be noted that the difference between the linear $\alpha_n$ and the constant $\alpha_n$ is only significant when the shift is small. When $f>0.1$, the difference is almost negligible.  

\subsection{AUCRSS vs E-AUCRSS}
\label{sec:EAUCRSS}
We evaluate whether the approximated adaptive sampling algorithm E-AUCRSS scarifies the detection power too much. The experiment parameter settings are the same as those in  Section \ref{sec:alpha}. To have a comprehensive understanding of detection power over different shift magnitudes, we set $ f = 0,\  0.1,\  ...,\  1.0$ .

\begin{figure*}[!h]
    \centering
    \subfloat[]{%
        \includegraphics[scale = 0.068]{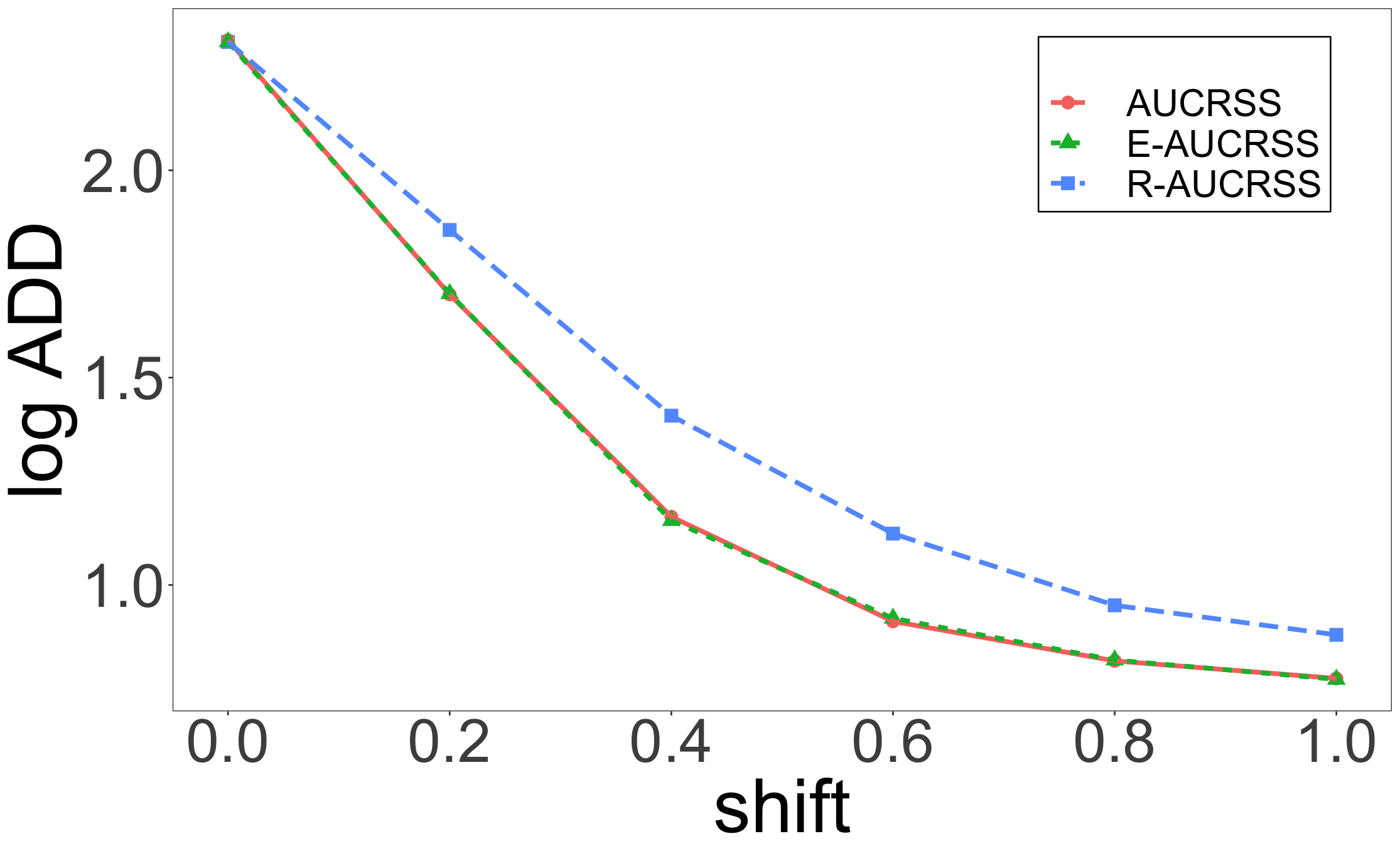}
        \label{fig:new_vs_originalm_2}
    }
    \subfloat[]{
        \includegraphics[scale = 0.068]{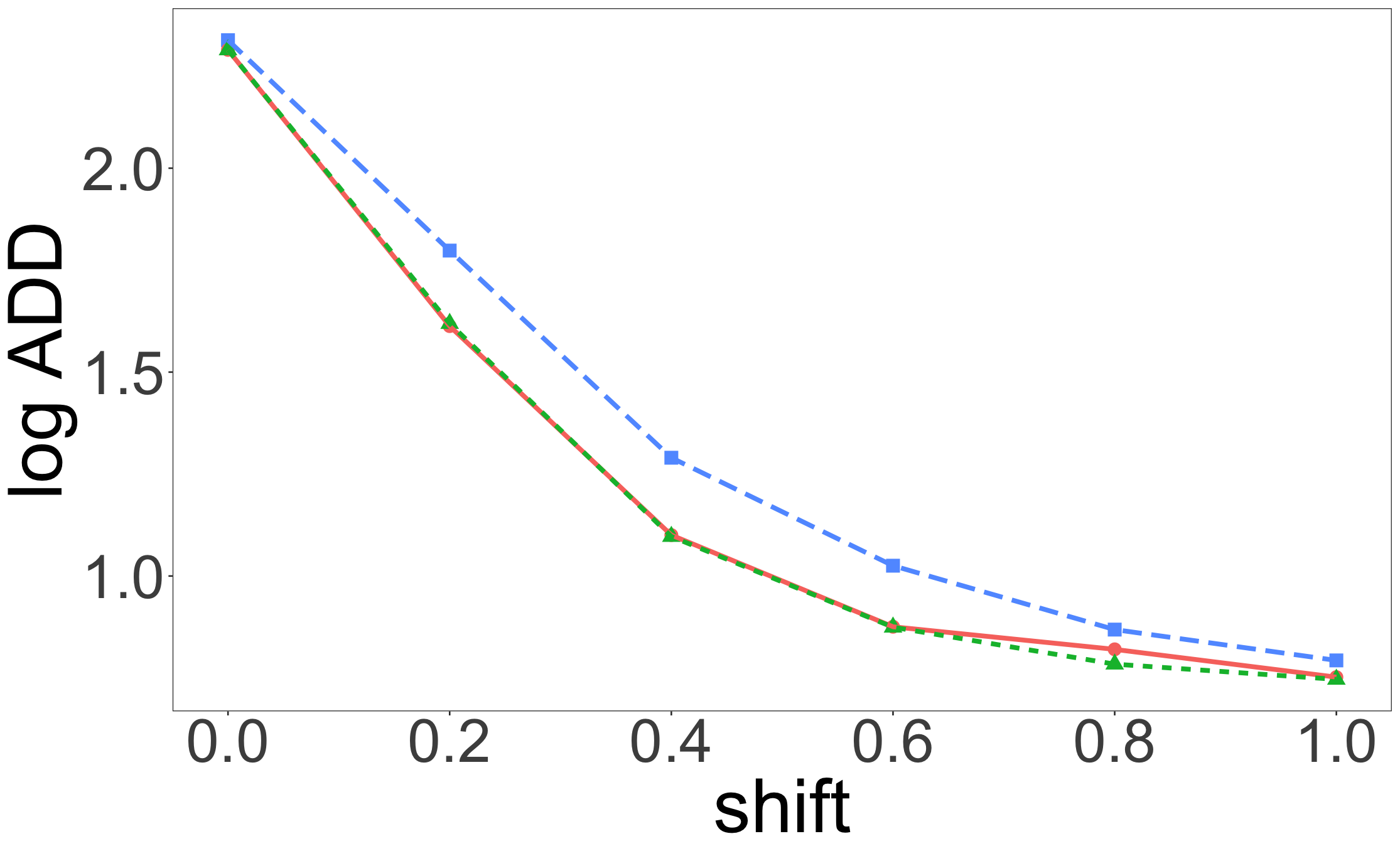}
        \label{fig:new_vs_originalm_3}
    }
    \subfloat[]{%
        \includegraphics[scale = 0.068]{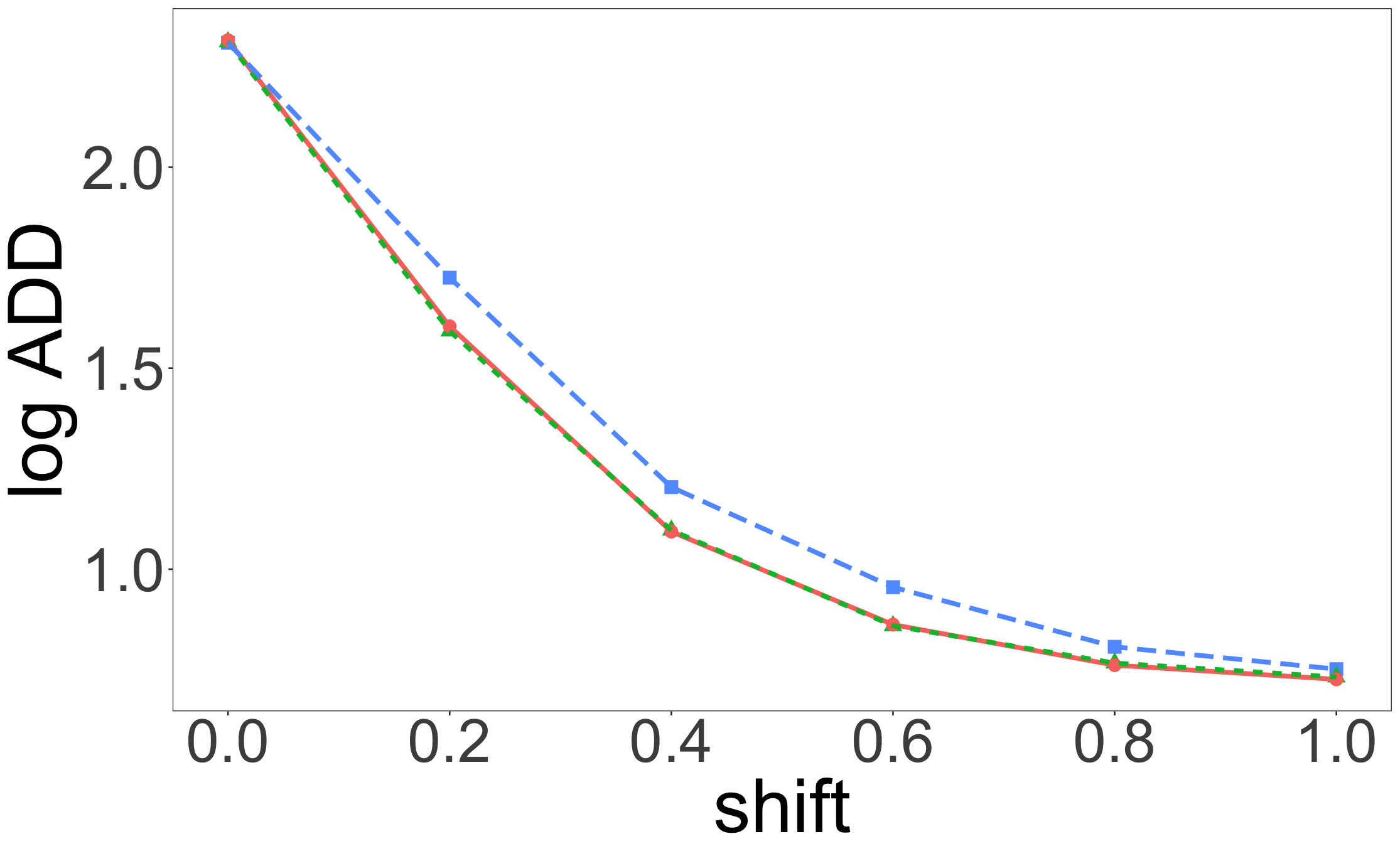}
        \label{fig:new_vs_originalm_4}
    }
    \caption{Comparison between AUCRSS and E-AUCRSS for $10$-dimensional data streams with (a) $m = 2$, (b) $m = 3$ and (c) $m = 4$}
    \label{fig:new_vs_original}
\end{figure*}

Note that E-AUCRSS and AUCRSS are equivalent when $m = 1$. Thus, in this section, we compare the performance of AUCRSS, E-AUCRSS and R-AUCRSS under $m = 2$ , $m = 3$ and $m = 4$, respectively.  The method for determining the threshold $h$ is the same as that in Section \ref{sec:alpha}, and we use the adaptive sampling strategy for both AUCRSS and E-AUCRSS. The results are shown in the Figure \ref{fig:new_vs_original}.

In Figure \ref{fig:new_vs_original}, we can find that both E-AUCRSS and AUCRSS perform much better than R-AUCRSS.  The difference between E-AUCRSS and AUCRSS is not significant. As we have mentioned, the complexity of E-AUCRSS is $O(mp)$, while that of AUCRSS is $O(C_p^m)$. Consider that for large-scale problems, E-AUCRSS takes much less computation than AUCRSS, but has almost same performance. When $m$ increases, the advantages of E-AUCRSS will become more prominent. As such, in the following experiments, we use E-AUCRSS for further extensive experiments and comparison with other benchmarks.

\begin{figure}[bht]
         
		\centering
		\subfloat[]{%
        \includegraphics[scale = 0.075]{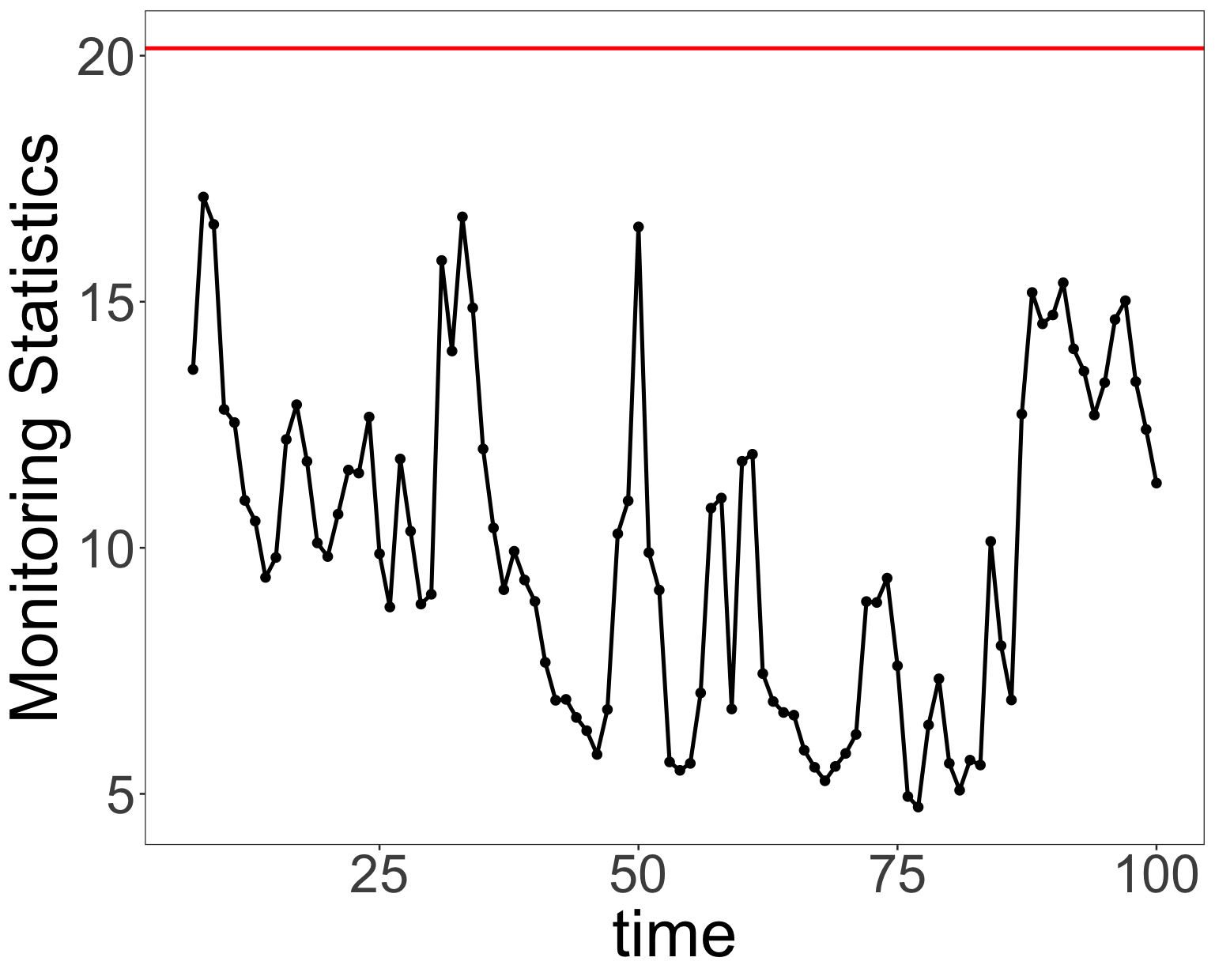}
        \label{fig:IC control}}
        \subfloat[]{%
        \includegraphics[scale = 0.075]{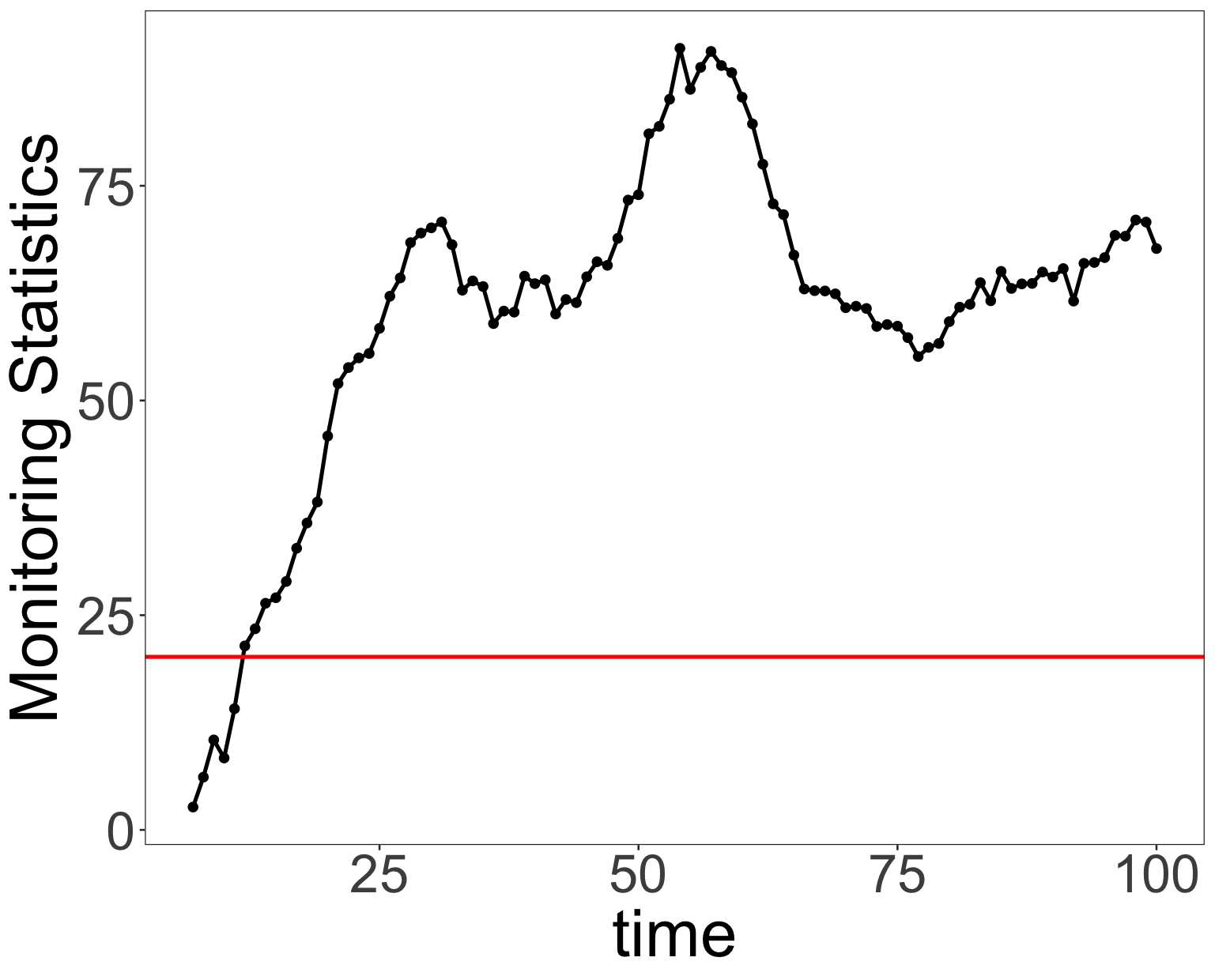}
        \label{fig:OC control}}
        \quad
        \subfloat[]{%
        \includegraphics[scale = 0.075]{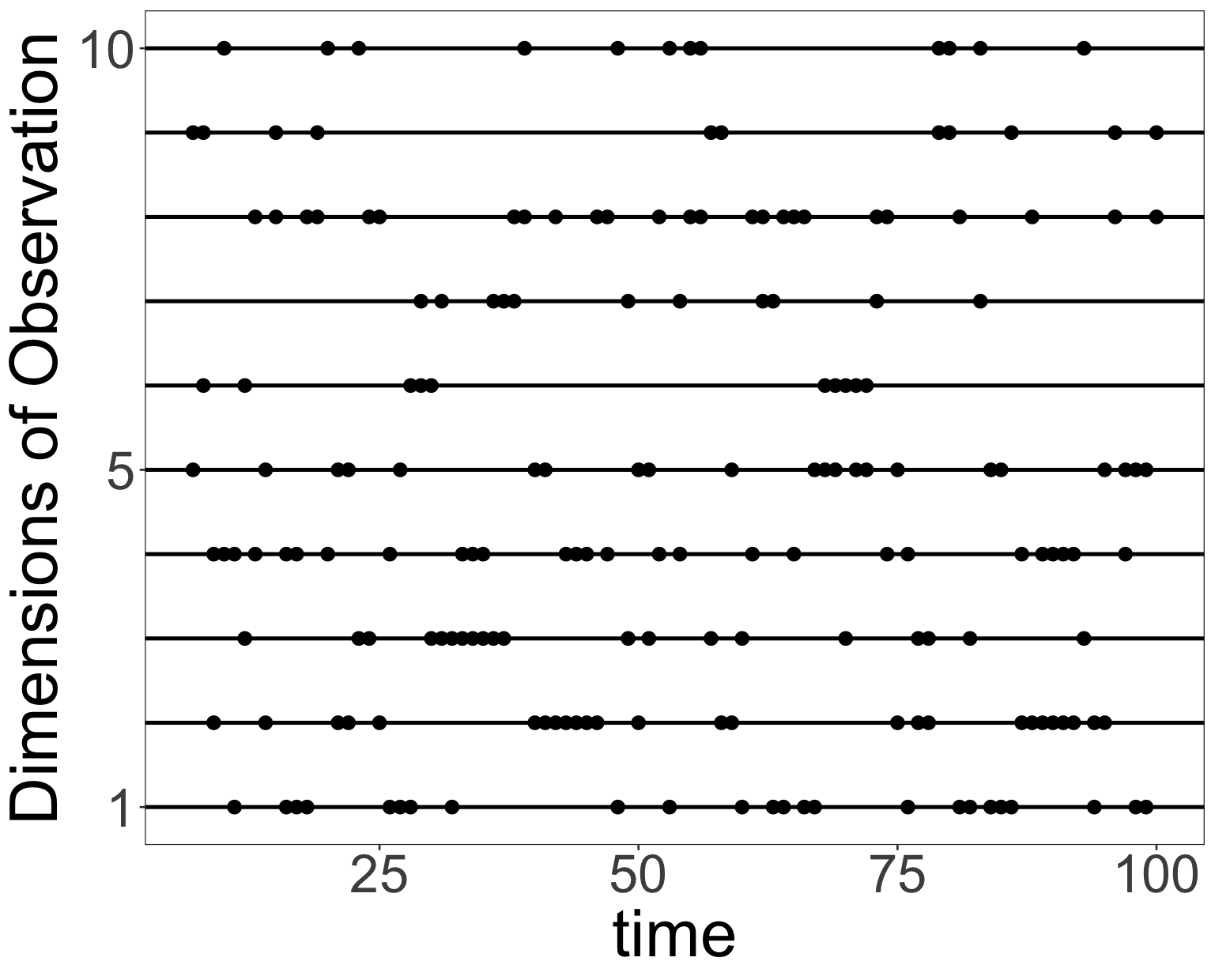}
        \label{fig:IC obse}}
        \subfloat[]{%
        \includegraphics[scale = 0.075]{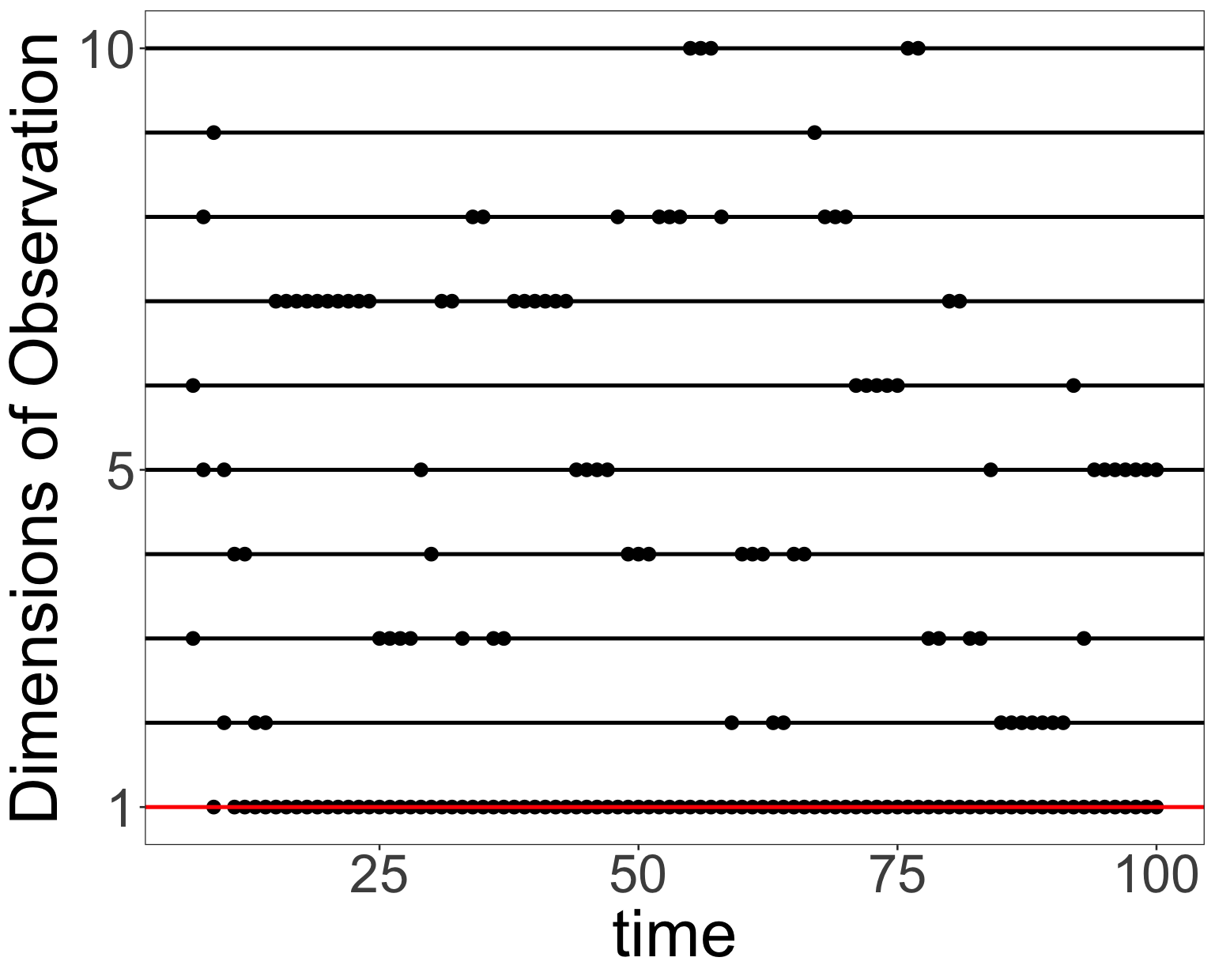}
        \label{fig:OC obse}}
    \caption{Online monitoring schematics for IC condition (a) and OC condition (b) and observe dimensions for IC condition (c) and OC condition (d) of E-AUCRSS}
    \label{fig:control chart}  		
\end{figure}

We also show the test statistics and the observed dimensions of E-AUCRSS under $m = 2$, $q = 10$ in Figure \ref{fig:control chart}. 
The test statistics and the observed dimensions for the first 100 time points of one IC run together with $h$ are shown in Figure \ref{fig:IC control} and Figure \ref{fig:IC obse}. The monitoring statistic remains relatively stationary below the control limit, and the observed dimensions are randomly selected as we proved in Theorem \ref{the:IC}. Then we further evaluate the OC performance. Similar as the IC case, we randomly select one OC run and observe its first 100 time points in Figure \ref{fig:OC control} and Figure \ref{fig:OC obse}. We can see that the monitoring statistic keeps increasing, and exceeds the control limit and triggers alarm efficiently. One observed dimension also fixes on the changed one highlighted in red. This is consistent with what we described in Lemma \ref{lemma_pt}, under OC condition, the algorithm will observe the changing dimensions with a high probability.

\subsection{Comparison with benchmarks}
\begin{figure}[htb]
    \centering
    \subfloat[]{%
        \includegraphics[scale = 0.07]{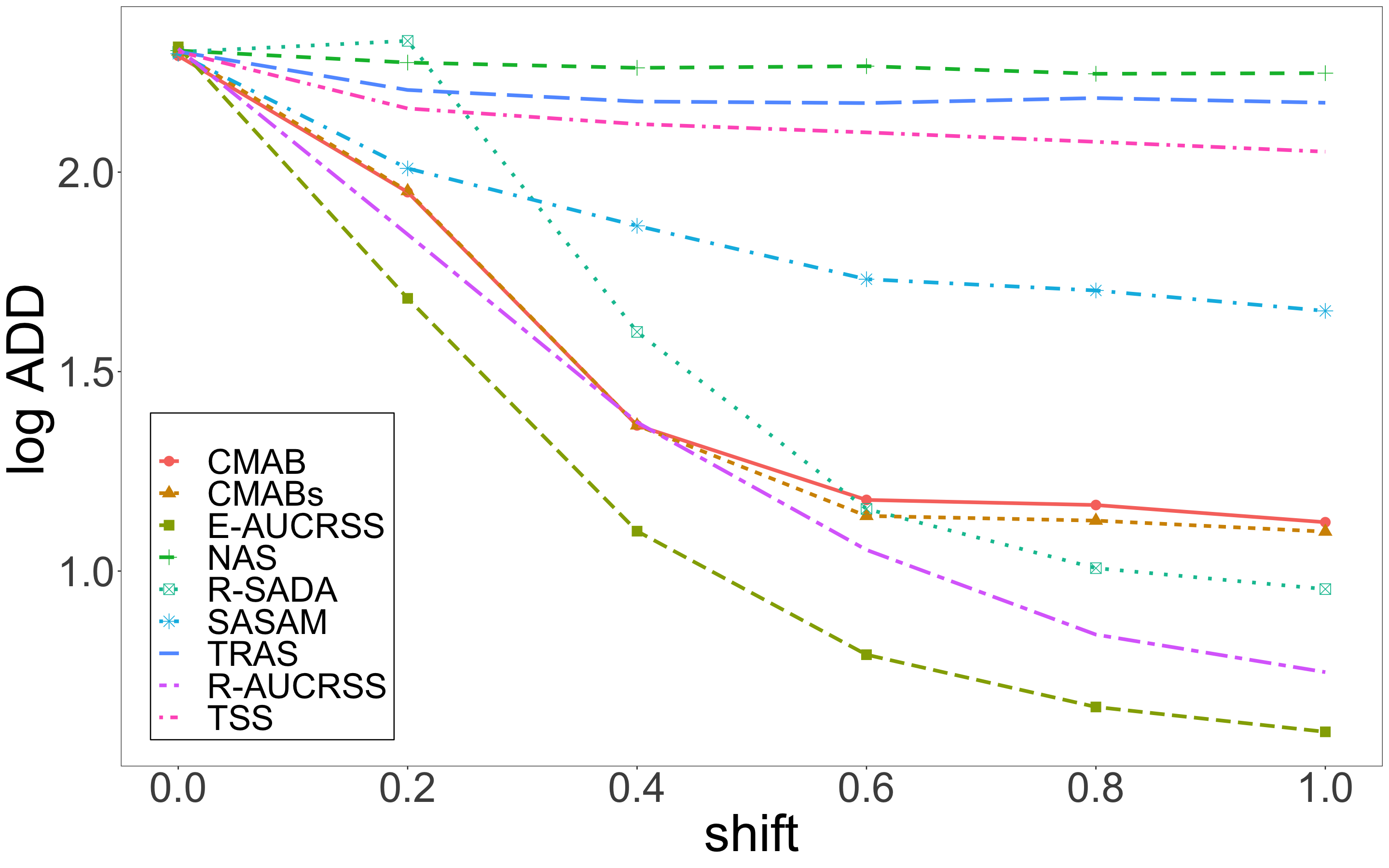}
        \label{fig:baselinep_10m_2}
    }
    \quad
    \subfloat[]{
        \includegraphics[scale = 0.07]{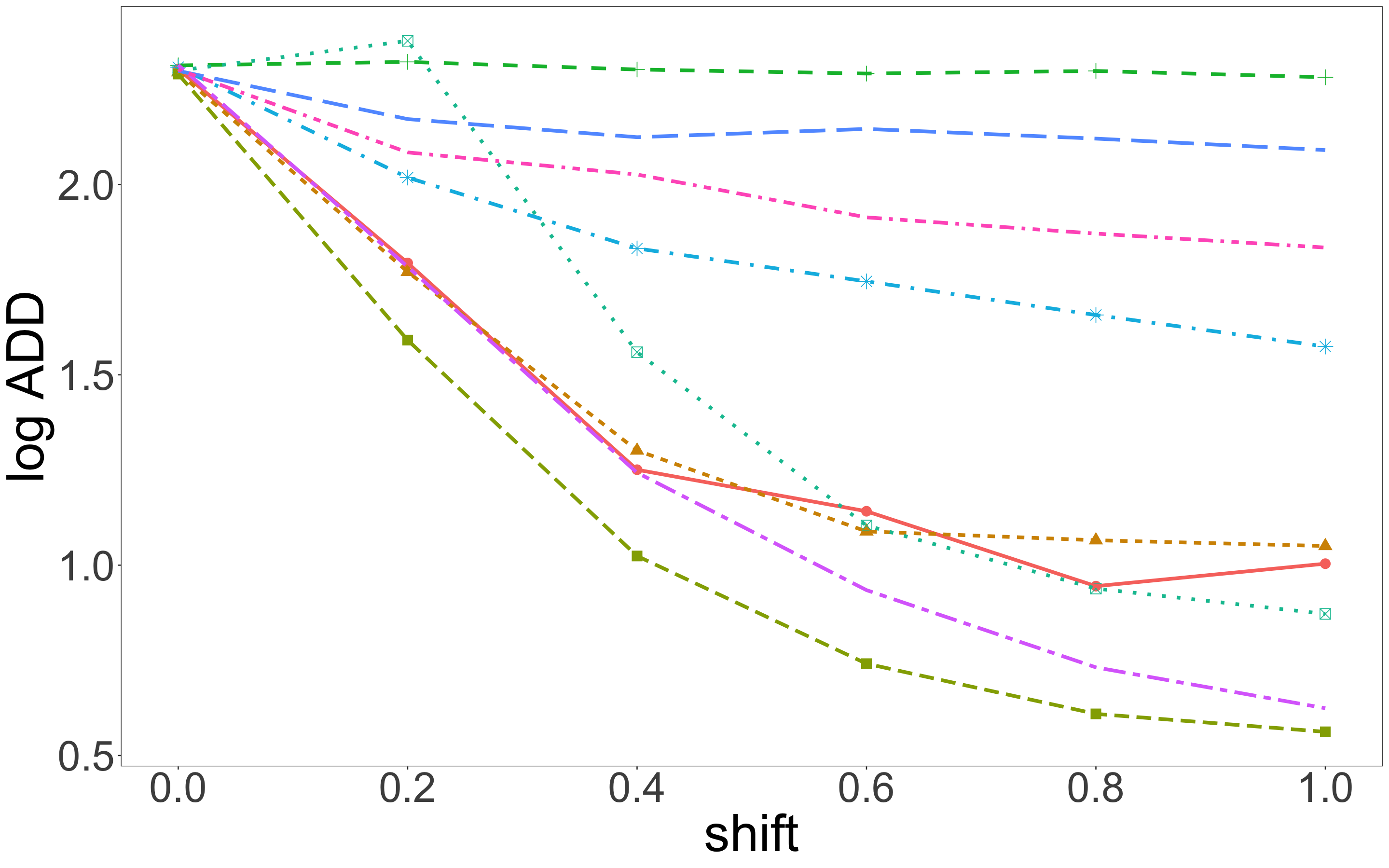}
        \label{fig:baselinep_10m_3}
    }
    \caption{Monitoring performance comparison for $10$-dimensional data streams with (a) $m=2$ and (b) $m=3$}
    \label{fig:baselinep_10}
\end{figure}

Now we compare E-AUCRSS with other benchmarks. The experiment parameter settings are the same as Section \ref{sec:EAUCRSS}. The detection results for different shift magnitudes with $m=2$ and $m=3$ are shown in Figure \ref{fig:baselinep_10}.

Clearly, E-AUCRSS has the best overall performance. Its advantage is extremely obvious for small $m$ and small change magnitudes. CMAB and CMABs perform secondly best because they lose the information of data autocorrelation which influences detection performance. R-SADA's performance is poor when the shift is small but becomes better as the shift increases.  On the one hand, this is because R-SADA does not consider the autocorrelation of data as well.  On the other hand, R-SADA has a threshold $k$. When the cumulative shift is less than $k$, the monitoring statistic will be reset to 0. In this case, it is difficult for R-SADA to detect the small change. In contrast, SASAM performs not so poorly when shift is small, but gets worse as the shift increases. This is mainly due to its CUSUM technique used in the monitoring statistic. As to NAS and TRAS, they perform unsatisfactorily for all the shifts' settings. The reason is that they ignore both the time autocorrelation and cross-correlation of the data. The performance of TSS is not satisfactory. This is because it assumes the data autocorrelation can vary over time smoothly in the IC model and aims to detect very abrupt changes that cannot be described by the time-variant IC model. Consequently, when the anomaly is not so abrupt as in our case, it is very likely to treat the anomaly as time-variant IC model's data and misdetect it. Furthermore, due to its heavy computational complexity of tensor operations, we can only use it for the small-scale simulation. Last, for all the methods, when the number of observable dimensions increases, the ADDs become smaller. This is expected since more observable dimensions can bring more information of the change point and lead to earlier detection.

\begin{figure}[htb]
    \centering
    \subfloat[]{%
        \includegraphics[scale = 0.07]{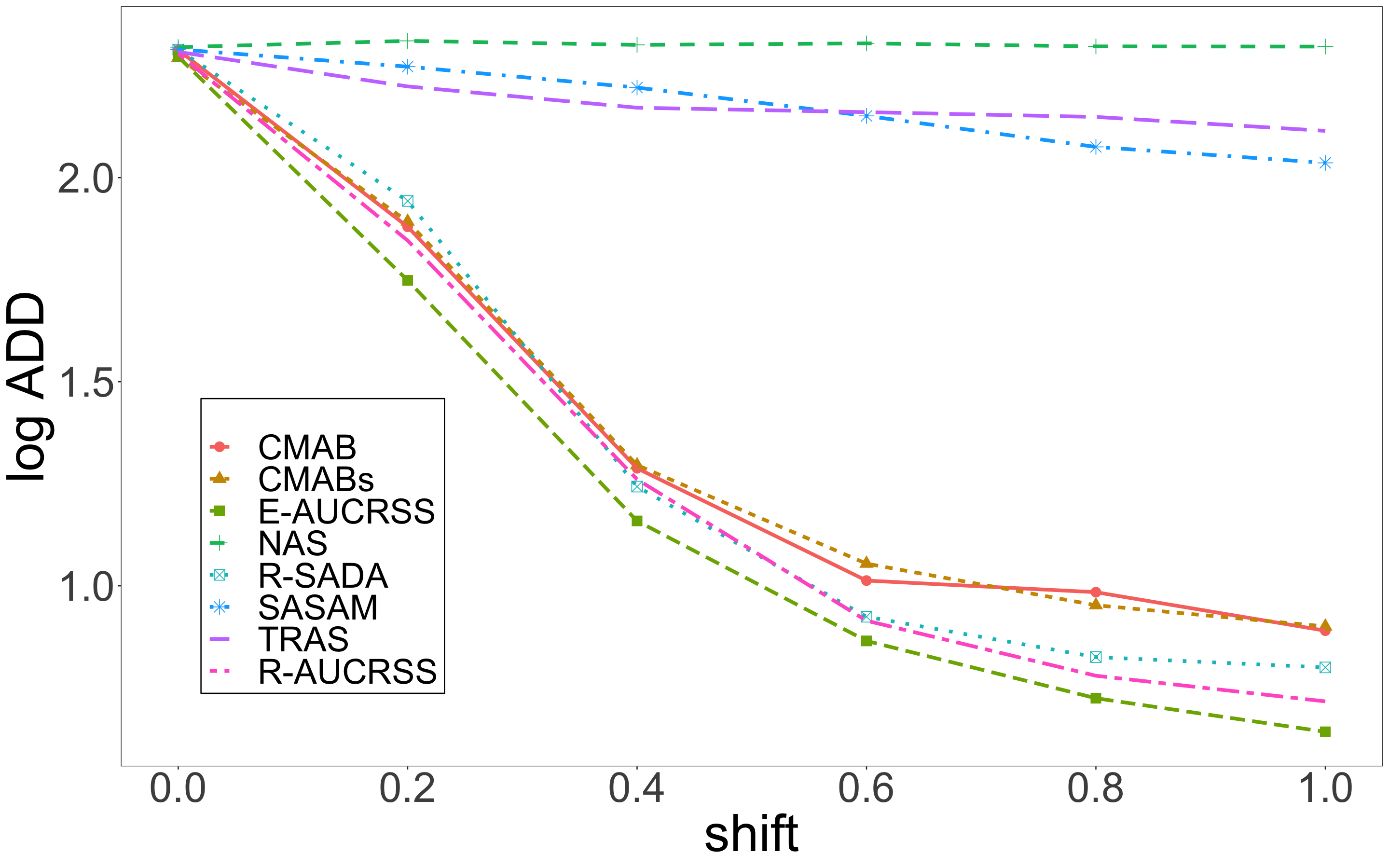}
        \label{fig:baselinep_30m_5}
    }
    \quad
    \subfloat[]{
        \includegraphics[scale = 0.07]{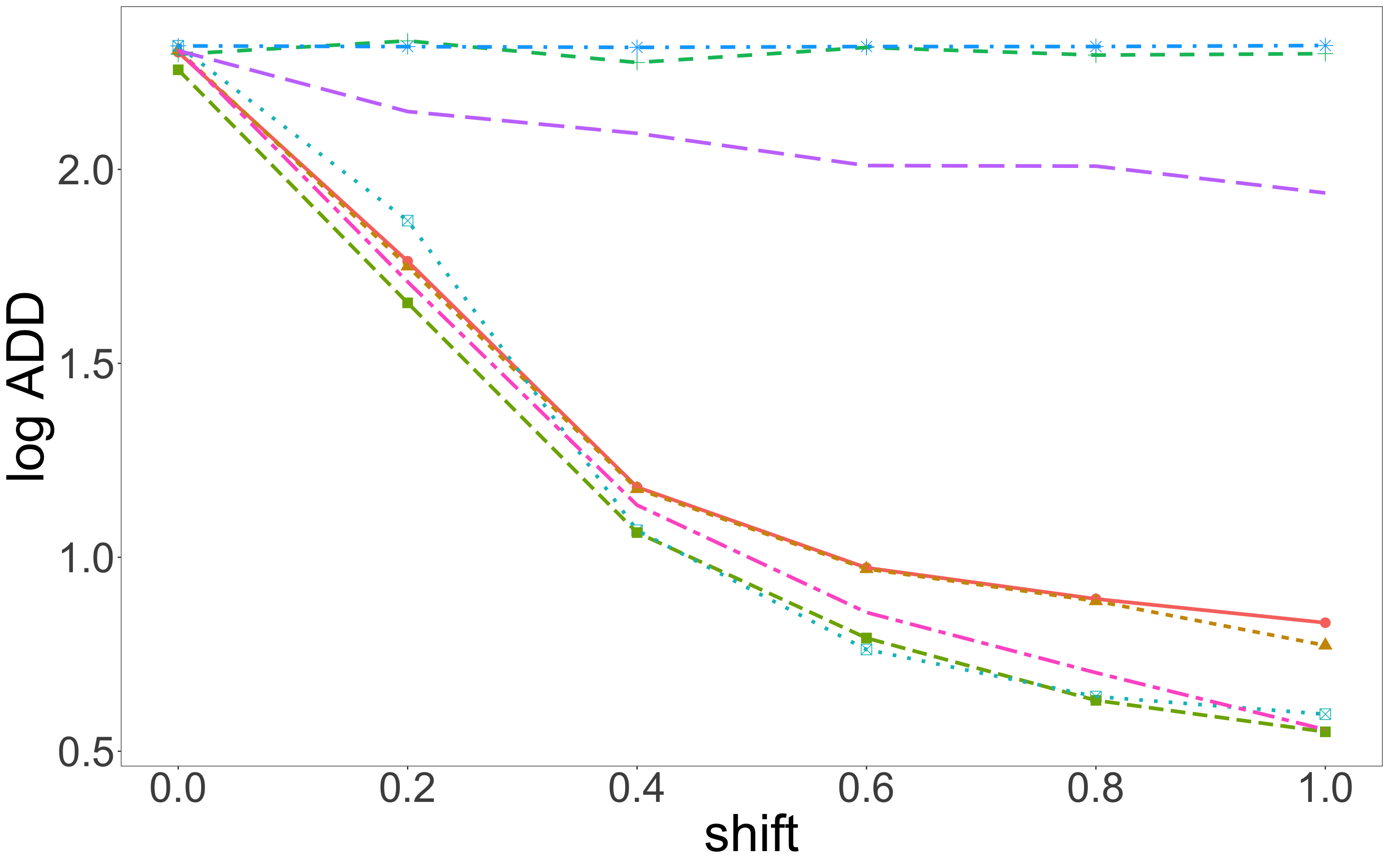}
        \label{fig:baselinep_30m_10}
    }
    \caption{Monitoring performance comparison for $30$-dimensional data streams with (a) $m=5$ and (b) $m=10$}
    \label{fig:baselinep_30}
\end{figure}

We also conduct experiments for high-dimensional cases with $p=30$ and $q = 15$. The experiment parameters keep the same as in  Section \ref{sec:EAUCRSS},  we set the transition matrix $\A$ as a sparse symmetric square matrix with diagonal elements 1 and eigenvalues less than 1. For the output matrix $\C$, we set its upper diagonal elements as 1 and set a non-zero element for each row to keep the sparsity. The detection results of different methods under $m=5$ and $m=10$ are shown in Figure \ref{fig:baselinep_30}. The performance comparison results are similar to those for $p=10$. Overall, E-AUCRSS performs the best, followed by CMAB and CMABs, which both perform almost equally well. R-SADA performs slightly worse than the three methods above, but still much better than NAS, SASAM,TSS and TRAS.

\section{Case Study}
\label{sec:case study}
In this section, we apply E-AUCRSS to a real data set of a milling process. The data are sampled by three different types of sensors (acoustic emission sensors, vibration sensors, and current sensors) with a total of six variables from ten different operating conditions of the machine. One condition can be treated as the IC condition, and the others can be treated as different OC conditions.  As we mentioned in Section \ref{sec:intro},  the data collected over time are autocorrelated and fit our model assumption well.

We briefly summarize the step-by-step implementation of E-AUCRSS as follows: 1) preprocessing with data normalization. 2) fitting data by a state space model; 3) applying E-AUCRSS to do monitoring under IC and OC conditions;. 4) comparing E-AUCRSS with other baselines.

There are 10 conditions. Under each condition, there are about 203 runs of data. Each run includes autocorrelated data collected from 500 sequential time points. We calculate the mean and standard deviation of each variable of data under IC condition to normalize all data. 
Then we fit the data by state space model using one run data of the IC condition, which is treated as historical off-line IC data.
We use EM algorithm to estimate the parameter $\bQ$, $\bR$, $\A$ and $\C$ in the state space model. It should be noted that we need to decide the dimension of state space $q$. Under different $q$, we use the estimated $\bQ$, $\bR$, $\A$ and $\C$ to calculate the mean square error (MSE) of SSM. We find that when $q$ reaches 10, further increasing  $q$ no longer significantly reduces the  MSE, so we finally select $q=10$. 

After estimating the parameter of the state space model, we apply the model to another run's data under IC condition. The model's one-step ahead prediction performance is shown in the appendix. 
It shows that the fitted state space model can predict the data well with a small prediction error. This demonstrates our state space model is very suitable for such data modeling. 

In this way, we can use E-AUCRSS for online monitoring. To increase the sample size, we need to perform uniform random sampling with replacement to transform 203 runs into 1000 runs. Then we assume that we can observe $m=2$ dimensions out of $p=6$ dimensions each time and set the time window $m_{1} = 50$ and $m_2 = 0$. We use the test statistics of IC condition's 1000 runs to calculate the control limit $h$ such that $ADD_{IC} = 200$. 
The specific selection method of $h$ is the same as Section V.A.

\begin{figure}
			\includegraphics[scale = 0.085]{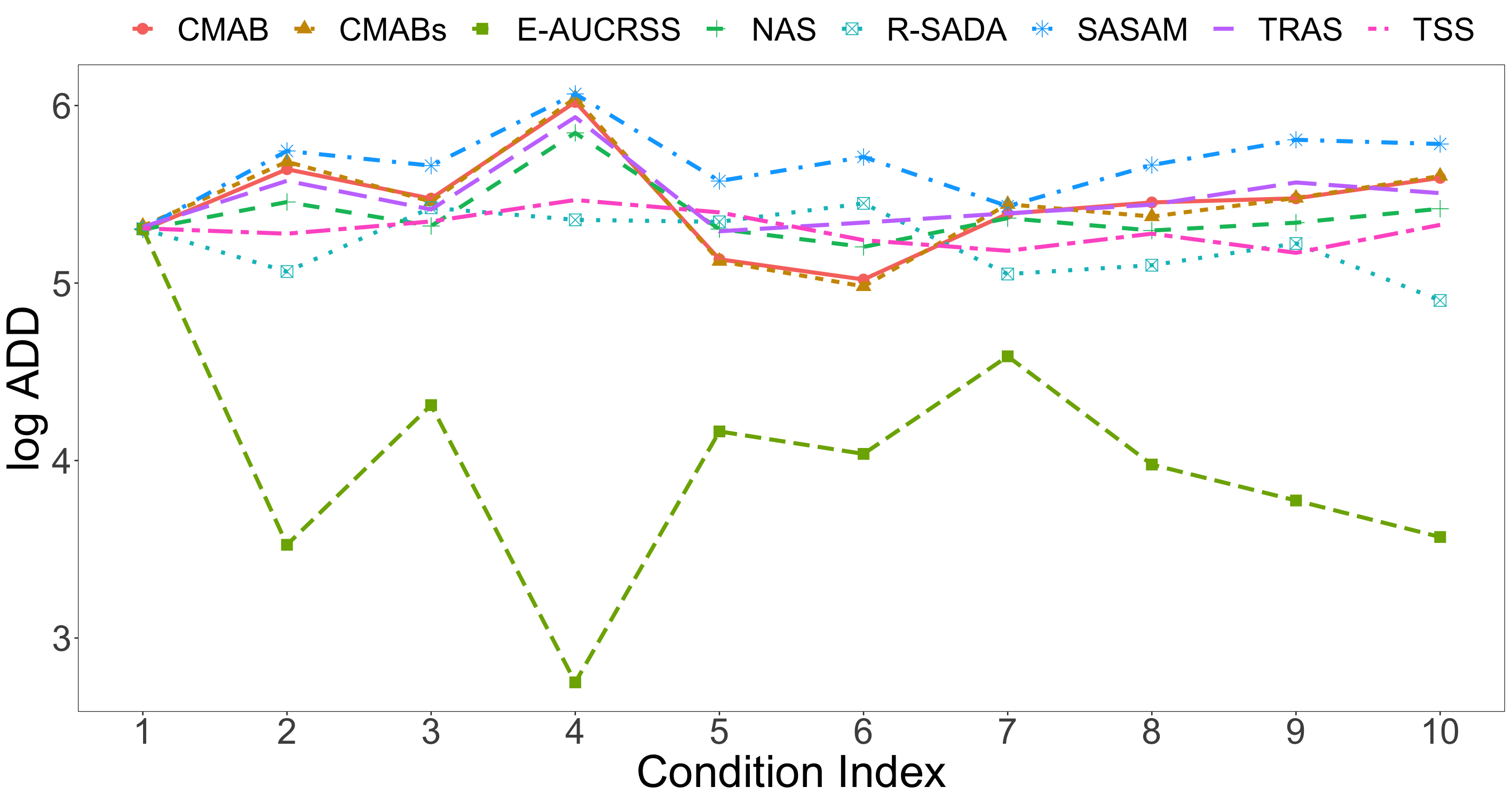}
			\caption{Monitoring performance comparison for the milling process data}
			\label{fig:case_baseline}
\end{figure}

To further have a comprehensive evaluation, we calculate the $ADD_{OC}$ for each OC condition, and compare E-AUCRSS with other benchmarks. The results are shown in Figure \ref{fig:case_baseline}. Clearly, our algorithm has a much smaller $ADD_{OC}$ than the other algorithms, which means that our algorithm has extremely excellent results for real-world application data with autocorrelation.  E-AUCRSS performs particularly well in condition four but other methods perform even worse in this condition. This is because the data in condition four have more stronger autocorrelation which can provide more information for our E-AUCRSS but violate the assumptions more seriously for other methods. Among the baselines, R-SADA performs better than the others, especially for condition four. This may be because R-SADA uses nonparametric test statistic, which can decrease the influence of data autocorrelation. Though NAS uses the same technique, it still performs unsatisfactorily, since it requires different dimensions of data to be uncorrelated. As to TSS, it has a bad performance due to the same reason as in numerical studies. 

\section{Conclusion}
\label{sec:conclusion}
Online monitoring for multivariate time series data is an extremely important topic. However, the condition with partially observable data has not been addressed so far. In this paper, we propose a monitoring scheme called AUCRSS. It uses a state space model to fit the multivariate time series data, where a partially-observable Kalman filter is proposed for model inference. The one-step ahead prediction error is used to construct the monitoring statistics by the generalized likelihood ratio test. Then by formulating the problem as a CMAB, an adaptive sampling strategy is proposed based on the upper-confidence region strategy. The detection power that is related to the sampling strategy is theoretically analyzed under  IC and OC condition respectively. Simulation studies and a real case study for milling process monitoring are conducted to demonstrate the advantages of the proposed method.

There are some topics to be further studied along this research detection. For example, the formulation of the adaptive $\alpha_n$ can be extended to nonlinear ones, like polynomial formulations. For the state space model, we can also try some new methods like nonparametric state space model to fit the system. In this way,  we may find a nonparametric monitoring statistic which is more robust. Last but not least, if the target data is high dimensional, we should try some dimension reduction methods for modeling and monitoring.

\bibliography{paper-ref}
\bibliographystyle{IEEEtran}

\end{document}